\documentclass[11pt]{article}
\PassOptionsToPackage{table}{xcolor}
\usepackage[final]{acl}

\usepackage{colortbl}     
\usepackage{booktabs}     
\usepackage{multirow}     
\usepackage{graphicx}     
\usepackage{amsmath}      
\usepackage[most]{tcolorbox}
\usepackage{setspace}

\usepackage{tabularx}
\usepackage[table]{xcolor}
\usepackage{CJKutf8}     
\usepackage{makecell}  

\definecolor{RecoverText}{RGB}{70,110,160} 

\definecolor{LighterBlue}{RGB}{240, 245, 255}
\definecolor{LighterGreen}{RGB}{242, 250, 242}
\definecolor{LighterOrange}{RGB}{255, 250, 240}
\definecolor{LighterPurple}{RGB}{250, 245, 255}
\definecolor{AclDarkBlue}{RGB}{33, 47, 61}
\definecolor{HeaderDark}{RGB}{24,54,88}     
\definecolor{GroupBlue}{RGB}{235,243,255}   
\definecolor{GroupOrange}{RGB}{255,244,233} 
\definecolor{OursPurple}{RGB}{243,239,255}  
\usepackage[table]{xcolor}
\definecolor{AclDarkBlue}{RGB}{24, 54, 88}

\newcommand{\zh}[1]{\begin{CJK*}{UTF8}{gbsn}#1\end{CJK*}}
\newcommand{\tw}[1]{\begin{CJK*}{UTF8}{bsmi}#1\end{CJK*}}
\newcommand{\ja}[1]{\begin{CJK}{UTF8}{min}#1\end{CJK}}
\newcommand{\ko}[1]{\begin{CJK}{UTF8}{mj}#1\end{CJK}}
\usepackage{xcolor,colortbl}
\definecolor{CaseBlue}{RGB}{28,78,128}
\definecolor{CaseBlueBg}{RGB}{235,244,255}
\definecolor{CaseGreen}{RGB}{22,120,74}
\definecolor{CaseGreenBg}{RGB}{232,248,239}
\definecolor{CaseRed}{RGB}{176,41,33}
\definecolor{CaseRedBg}{RGB}{253,236,234}
\definecolor{CaseGray}{RGB}{90,90,90}

\definecolor{MainBlue}{RGB}{45,85,125}
\definecolor{WarnRed}{RGB}{150,50,45}
\definecolor{SoftGray}{RGB}{245,247,249}
\definecolor{LineGray}{RGB}{160,160,160}
\definecolor{HSSGray}{RGB}{242,244,247}

\definecolor{HSSLightBlue}{RGB}{46,134,193}

\definecolor{HSSRowGray}{RGB}{248,249,251}

\usepackage{booktabs,tabularx,multirow,xcolor,colortbl}
\usepackage{siunitx}

\definecolor{LighterGray}{gray}{0.965}           
\definecolor{LightGold}{RGB}{255,253,240}        
\definecolor{GoldLine}{RGB}{235,224,190}         

\newcommand{\labeltag}[2]{%
  \tcbox[
    on line,
    colback=white,
    colframe=#1,
    boxrule=0.5pt,
    arc=0.8mm,
    left=3pt,right=3pt,top=1pt,bottom=1pt
  ]{\textbf{\footnotesize\textcolor{#1}{#2}}}%
}
\tcbset{
  compactbox/.style={
    before skip=0pt,
    after skip=0pt,
    boxsep=0pt
  }
}

\usepackage{times}
\usepackage{latexsym}

\usepackage[T1]{fontenc}

\usepackage[utf8]{inputenc}

\usepackage{microtype}

\usepackage{inconsolata}

\usepackage{graphicx}

\usepackage{booktabs, tabularx, multirow, multicol, makecell, colortbl}

\usepackage{enumitem}

\usepackage{bbding}

\usepackage{amsmath, amssymb}

\usepackage{marvosym}
\usepackage{float}
\usepackage{footmisc}

\usepackage[table]{xcolor}

\title{How Order-Sensitive Are LLMs? \\
OrderProbe for Deterministic Structural Reconstruction}

\newcommand{\equal}{\textsuperscript{*}}
\newcommand{\corr}{\textsuperscript{\dag}}

\author{
\textbf{Zhaolu Kang}\equal\textsuperscript{1,2},
\textbf{Yingjie He}\equal\textsuperscript{2},\\
\textbf{Kehan Jiang}\textsuperscript{2},
\textbf{Leqi Zheng}\textsuperscript{3},
\textbf{Jiachen Qian}\textsuperscript{4},
\textbf{Qianyuan Zhang}\textsuperscript{5},
\textbf{Chunlei Meng}\textsuperscript{6},\\
\textbf{Yujie Feng}\textsuperscript{7},
\textbf{Yuan Wang}\textsuperscript{8},
\textbf{Jiabao Dou},
\textbf{Aming Wu},
\textbf{Pengxiang Zhao}\textsuperscript{8},\\
\textbf{Jiaxin Liu}\textsuperscript{9},
\textbf{Guansu Wang}\textsuperscript{10},
\textbf{Zeyu Zhang},
\textbf{Lei Wang}\textsuperscript{2},
\textbf{Qishi Zhan}\textsuperscript{11},\\
\textbf{Xiaomin He}\textsuperscript{2},
\textbf{Meisheng Zhang}\textsuperscript{2},
\textbf{Jianyuan Ni},
\textbf{Richeng Xuan}\corr\textsuperscript{1}
\\[2pt]
\textsuperscript{1}Hunyuan, Tencent,   
\textsuperscript{2}Peking University\\
\textsuperscript{3}Tsinghua University,
\textsuperscript{4}City University of Hong Kong,
\textsuperscript{5}The Chinese University of Hong Kong,\\
\textsuperscript{6}Fudan University,
\textsuperscript{7}The Hong Kong Polytechnic University,
\textsuperscript{8}Zhejiang University,
\\
\textsuperscript{9}University of Illinois Urbana-Champaign,
\textsuperscript{10}Yale University,
\textsuperscript{11}Marquette University\\
\texttt{\selectfont{zlkang25}@stu.pku.edu.cn}
}

\usepackage{booktabs} 
\usepackage{amssymb}  
\usepackage{pifont}   
\usepackage{makecell} 

\newcommand{\cmark}{\ding{51}}%
\newcommand{\xmark}{\ding{55}}%

\begin{document}
\maketitle

\begingroup
  \renewcommand\thefootnote{\fnsymbol{footnote}}
  \footnotetext[1]{Equal contribution.}
  \footnotetext[2]{Corresponding author.}
\endgroup

\begin{abstract}
Large language models (LLMs) excel at semantic understanding, yet their ability to reconstruct internal structure from scrambled inputs remains underexplored. Sentence-level restoration is difficult to evaluate automatically because scrambled sentences often admit multiple valid reorderings. We introduce \textbf{OrderProbe}, a deterministic benchmark for structural reconstruction using fixed four-character expressions in Chinese, Japanese, and Korean, which have a unique canonical order and thus support exact-match scoring. We further propose a diagnostic framework that evaluates models beyond recovery accuracy, including Semantic Accuracy, Logical Validity, Structural Consistency, Robustness, and Information Density. Experiments on twelve widely used LLMs show that structural reconstruction remains difficult even for frontier systems: zero-shot recovery frequently falls below 35\%. We also observe a consistent gap between meaning-oriented generation and exact structural reconstruction, suggesting that structural robustness is not an automatic byproduct of semantic competence. 
To facilitate reproducibility, we publicly release OrderProbe and its evaluation code.
\end{abstract}

\begin{figure}[t]
    \centering
    \includegraphics[width=0.92\linewidth]{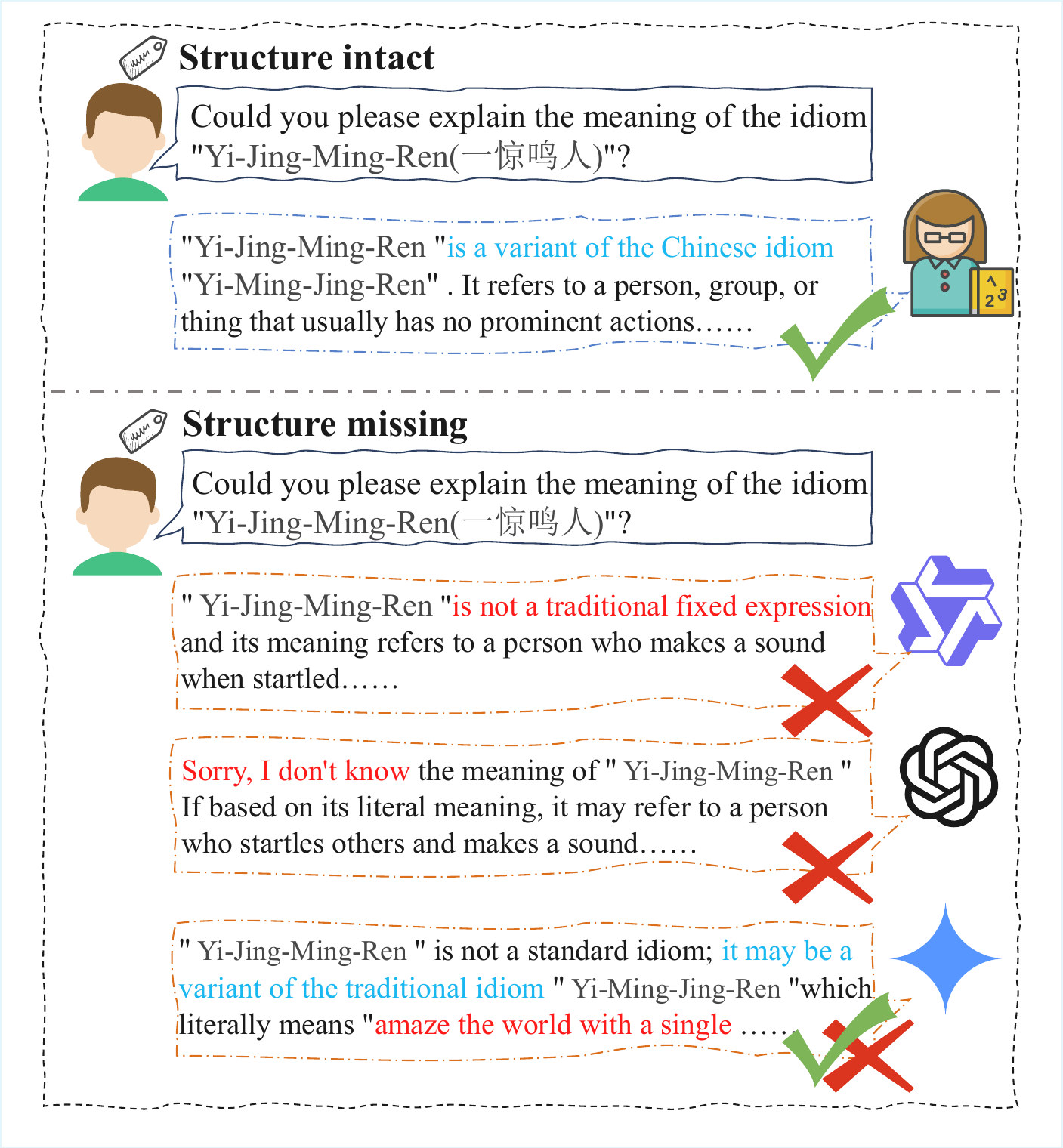}
    \vspace{-6pt}
    \caption{\textbf{OrderProbe example.} Humans recover the canonical order using semantic anchors, while LLMs often generate correct meanings but fail exact reconstruction.}

    \label{fig:orderprobe_example}
\vspace{-0.3cm}
\end{figure}

\definecolor{GroupGray}{gray}{0.94}
\definecolor{IdiomGray}{RGB}{248,246,242}

\begin{table*}[t!]
\centering
\small
\setlength{\tabcolsep}{6pt}
\renewcommand{\arraystretch}{1.15}

\resizebox{\textwidth}{!}{%
\begin{tabular}{l|c|c|c|c|c}
\toprule
\rowcolor{gray!15}
\textbf{Benchmark} & 
\textbf{Year} & 
\textbf{Language} & 
\textbf{Target Unit} & 
\textbf{Perturbation Level} & 
\textbf{Core Task} \\ 
\midrule

\rowcolor{GroupGray}
\multicolumn{6}{l}{\textbf{General Robustness Benchmarks}} \\

RoTBench \citep{ye2024rotbenchmultilevelbenchmarkevaluating} 
& 2024 & En & Tool-use I/O & Real-world Noise & Tool-use Robustness (Tool Learning) \\

Instruct-Robust \citep{li-etal-2024-evaluating-instruction} 
& 2024 & En & Instructions & External Injection & Instruction-Following Robustness \\

ReDial \citep{lin-etal-2025-assessing} 
& 2025 & En (Dialect) & Logic Problems & Dialectal Variation & Reasoning (Math/Logic) \\

\midrule

\rowcolor{IdiomGray}
\multicolumn{6}{l}{\textbf{Idiom \& Figurative Language Benchmarks}} \\

MMFLD \citep{lai-etal-2023-multilingual} 
& 2023 & Multi & Fig. Language & \textit{None} (Static) & Detection / Class. \\

WenMind \citep{NEURIPS2024_5c1019b5} 
& 2024 & Zh & Classical Text & \textit{None} (Static) & QA \& Generation \\

Chengyu-Bench \citep{fu2025chengyubenchbenchmarkinglargelanguage} 
& 2025 & Zh & Idioms & \textit{None} (Static) & Understanding \& Use \\

MChIRC \citep{WangWSSHHS25} 
& 2025 & Zh & Idiom-in-Context & \textit{None} (Static) & Multimodal RC / Cloze \\

Idiom-Trans \citep{donthi-etal-2025-improving} 
& 2025 & Multi & Idioms & \textit{None} (Static) & Translation \\

\midrule

\rowcolor{gray!8}
\textbf{OrderProbe (Ours)} 
& \textbf{2025} & \textbf{CJK} & \textbf{Idioms} & \textbf{Internal Structure} & \textbf{Reconstruction} \\

\bottomrule
\end{tabular}%
}

\vspace{-6pt}
\caption{\textbf{Positioning of OrderProbe.} Existing robustness benchmarks mainly test surface-level noise or instruction-level perturbations, while idiom benchmarks treat expressions as static retrieval or transformation targets. OrderProbe uniquely evaluates deterministic internal-order reconstruction with exact-match scoring.}
\label{tab:comparison}
\vspace{-0.3 cm}
\end{table*}

\section{Introduction}

Large language models (LLMs) have demonstrated strong performance in semantic understanding and text generation, yet their robustness to disruptions in internal structure remains unclear \citep{openai2024gpt4technicalreport, deepseekai2025deepseekv3technicalreport}. Humans can often recover meaning from structurally disordered inputs with little effort, but evaluating this ability in machine models is difficult. At the sentence level, restoration is inherently ambiguous because a scrambled sentence may admit multiple valid reorderings. This makes deterministic evaluation difficult and blurs the distinction between structural competence and general fluency.

Existing robustness benchmarks mainly focus on surface noise or external perturbations such as instruction manipulation \citep{wang-etal-2021-textflint,ribeiro-etal-2020-beyond,li-etal-2024-evaluating-instruction}. Although these benchmarks have advanced robustness evaluation, they do not directly test whether a model can reconstruct a uniquely constrained sequence from scrambled components. This limitation is particularly relevant given evidence that Transformer models can exhibit limited sensitivity to word order \citep{sinha-etal-2021-unnatural} and may fail to generalize structural relations under reversed or reordered contexts \citep{berglund2024the}.

To enable deterministic evaluation, we adopt fixed four-character expressions in Chinese, Japanese, and Korean as the target unit. Unlike scrambled sentences, such expressions typically have a single conventional order, making exact-match evaluation possible. We therefore study \textit{structural reconstruction}: recovering the canonical internal order of lexicalized expressions under controlled permutations, rather than interpreting arbitrary character combinations. This setting differs from existing idiom and figurative language benchmarks, which mainly evaluate retrieval, interpretation, translation, or usage under canonical inputs \citep{zheng-etal-2019-chid, lai-etal-2023-multilingual, fu2025chengyubenchbenchmarkinglargelanguage, donthi-etal-2025-improving, NEURIPS2024_5c1019b5}.

Figure \ref{fig:orderprobe_example} illustrates the core challenge. Humans can often recover the canonical order from scrambled characters, whereas LLMs frequently produce meaning-aligned explanations while failing exact reconstruction. This reveals a gap between semantic access and exact structural recovery.

To study this phenomenon systematically, we introduce \textbf{OrderProbe}, a benchmark for evaluating structural reconstruction in Chinese, Japanese, and Korean. The dataset is built through multi-source collection, expert filtering, and hybrid semantic annotation to ensure linguistic reliability. It contains 3,543 curated expressions spanning six syntactic categories and multiple script types. We further propose a diagnostic evaluation framework that goes beyond recovery accuracy by measuring Semantic Accuracy, Logical Validity, Structural Consistency, Robustness, and Information Density. This enables fine-grained analysis of failure patterns rather than simple leaderboard comparison.

Our contributions are summarized as follows.
\begin{itemize}[leftmargin=*,itemsep=1pt,topsep=2pt]
\item We identify the ambiguity of sentence-level restoration and propose fixed four-character expressions as a deterministic proxy for evaluating structural reconstruction.
\vspace{-6pt}
\item We introduce OrderProbe, a multilingual benchmark containing 3,543 curated expressions across six syntactic categories.
\vspace{-6pt}
\item We propose a diagnostic framework that analyzes Robustness, Semantic Accuracy, and generation errors beyond simple accuracy metrics.
\vspace{-6pt}
\item We evaluate twelve widely used LLMs on OrderProbe and show that exact reconstruction remains difficult, with zero-shot recovery often below 35\% even for strong systems.
\end{itemize}

\begin{figure*}[t!]
    \centering
    \includegraphics[width=1\linewidth]{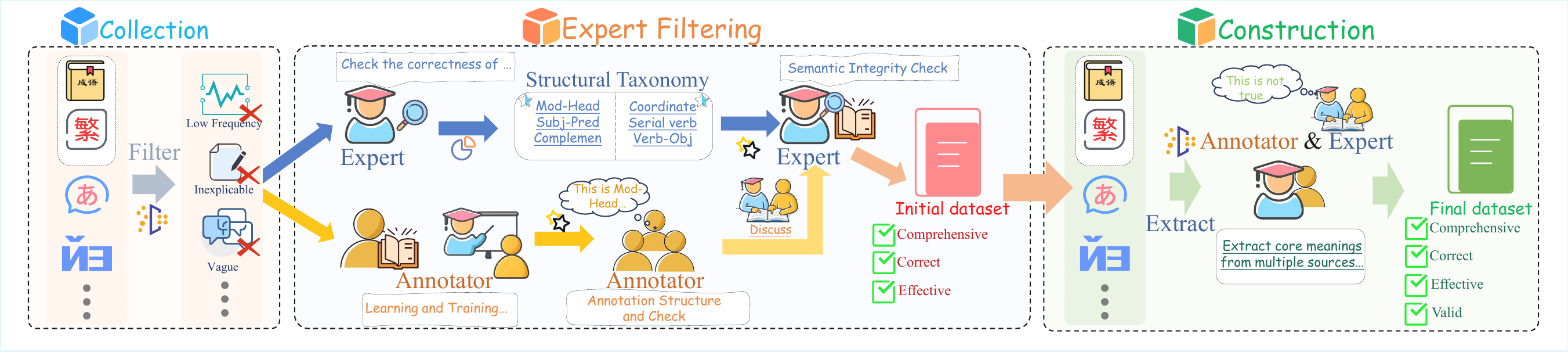}
    \vspace{-12pt}
    \caption{\textbf{OrderProbe construction pipeline.} We collect raw four-character expressions from multiple sources, apply expert filtering to obtain canonical roots, build semantic references via dictionary grounding and verified augmentation, and generate the full non-identity permutation space as structurally perturbed evaluation inputs.}
    \label{fig:construction_pipeline}
    \vspace{-0.3 cm}
\end{figure*}

\section{OrderProbe}
\label{sec:orderprobe}

OrderProbe is constructed through a controlled pipeline that transforms raw lexical resources into a deterministic benchmark for structural reconstruction. 
The pipeline consists of four stages: multi-source collection, expert filtering, semantic reference construction, and permutation-based perturbation generation. 
This section describes each stage and summarizes the resulting dataset.

\subsection{Stage I: Multi-source Collection}
We first assemble a candidate pool $\mathcal{C}_{raw}$ by aggregating four-character expressions from multiple publicly available resources, including lexical dictionaries and digital repositories. 
Each candidate item is associated with a script typology $T \in \{\textsc{ZH-CN}, \textsc{ZH-TW}, \textsc{JA}, \textsc{KO}\}$. 
The goal of this stage is to maximize coverage and diversity before quality control. 
This process yields approximately 4,000 raw candidates.

\subsection{Stage II: Expert Filtering}
To obtain a linguistically reliable core set, we apply an expert filtering function $\mathcal{F}$ that maps $\mathcal{C}_{raw}$ to a refined set $\mathcal{X}$:
\begin{equation}
\mathcal{X} = \mathcal{F}(\mathcal{C}_{raw}).
\end{equation}
Filtering is performed by senior linguists and trained annotators with proficiency in the target languages.
Annotations are conducted independently by five native-speaker annotators for each language, and final decisions are made by majority vote, followed by a brief consensus review for disputed cases.
Items are removed if they are non-standard, ambiguous, modern internet slang, or exhibit multiple competing canonical forms.
Expressions whose permutations could yield another conventional lexicalized entry are also removed to ensure a unique canonical target under the reconstruction objective.
Inter-annotator agreement reaches substantial consistency (Fleiss' $\kappa = 0.835$).
This process yields exactly 3,543 canonical expressions in $\mathcal{X}$, which form the benchmark roots.

\subsection{Stage III: Semantic Reference Construction}
For each canonical expression $x \in \mathcal{X}$, we construct a semantic reference set $\mathcal{S}_x$ to support evaluation beyond exact recovery. 
We retrieve dictionary definitions $d_{dict}$ as the primary semantic anchor, following common practice in idiom evaluation \citep{zheng-etal-2019-chid, fu2025chengyubenchbenchmarkinglargelanguage}. 
To reduce brittleness to surface phrasing, we augment each definition with paraphrastic variants generated by LLMs and verified by annotators, similar to recent benchmark construction protocols \citep{li-etal-2024-evaluating-instruction}. 
The final reference set is defined as:
\begin{equation}
\mathcal{S}_x = \{d_{dict}\} \cup \mathcal{D}_{aug}.
\end{equation}
This hybrid construction enables robust semantic and logical evaluation without translating across languages.

\subsection{Stage IV: Permutation-based Perturbation Generation}
The benchmark evaluates whether models can reconstruct the canonical order of an expression from scrambled constituents. 
Each canonical expression $x$ is represented as a token sequence $(t_1, t_2, t_3, t_4)$ with a fixed order. 
We apply a perturbation operator $\mathcal{P}$ that generates all non-identity permutations:
\begin{equation}
\mathcal{P}(x) = \{\pi(x) \mid \pi \in S_4, \pi(x) \neq x\},
\end{equation}
where $S_4$ denotes the permutation group over four elements. 
This yields 23 perturbed inputs per canonical expression, forming a complete combinatorial space for internal reordering. 
These permutations are controlled reconstruction stimuli rather than intended natural variants: they preserve the same four constituent units while disrupting their canonical internal order and retaining a unique gold target.

\subsection{Dataset Statistics}
\begin{figure}[!t]
    \centering
    \includegraphics[width=1\linewidth]{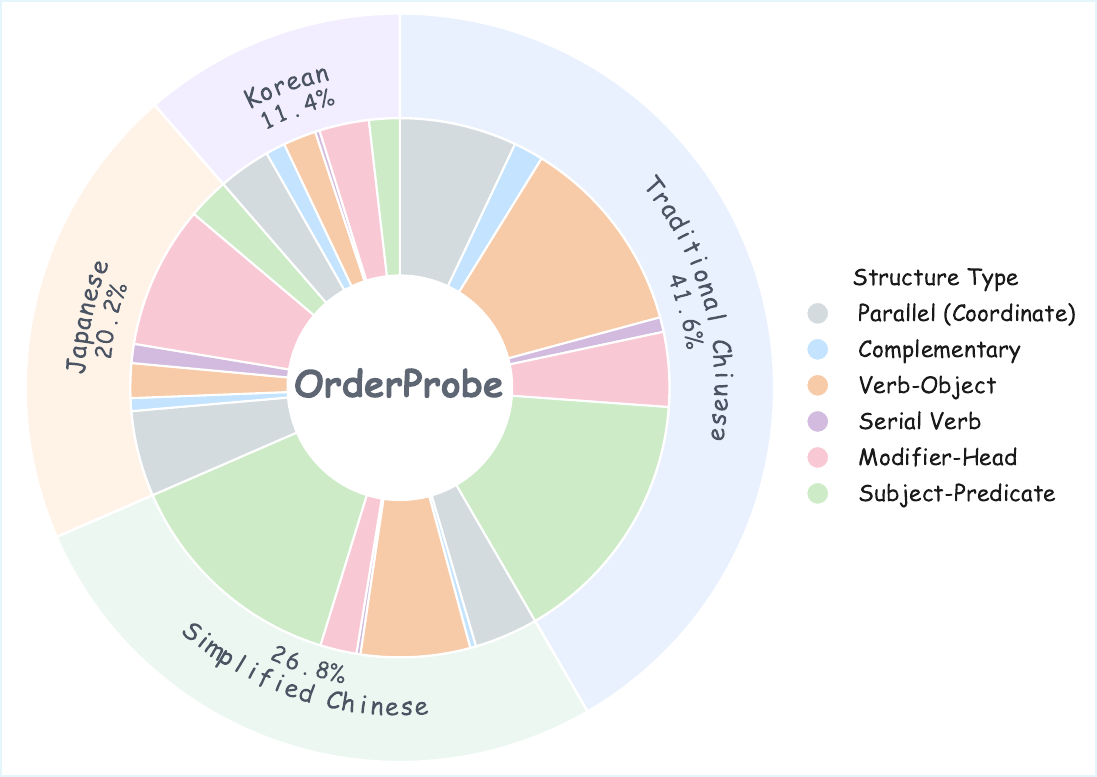}
    \vspace{-6pt}
    \caption{\textbf{Overview of OrderProbe.} The chart displays the distribution of script typologies and syntactic structures across the 3543 samples.}
    \label{figdataset}
    \vspace{-0.1cm}
\end{figure}

The final dataset $\mathcal{D}$ consists of all perturbed variants generated from the canonical set $\mathcal{X}$. 
It covers four script settings (Simplified Chinese, Traditional Chinese, Japanese, and Korean) and six syntactic categories. 
The distribution of script typologies and structural patterns is shown in Figure~\ref{figdataset}. 
This design supports controlled analyses of how reconstruction varies across writing systems and syntactic templates. We treat Korean Hangul as a phonogrammatic control: syllable-block scrambling preserves syllable composition but provides weaker standalone semantic anchors than logographic scripts.

\begin{table}[t!]
\centering
\small

\newcommand{\CatW}{1.6cm}
\newcommand{\MetW}{1.25cm}

\setlength{\tabcolsep}{1.6pt}

\renewcommand{\arraystretch}{1.12}

\setlength{\aboverulesep}{0pt}
\setlength{\belowrulesep}{0pt}

\begin{tabularx}{\linewidth}{p{\CatW} p{\MetW} X}
\toprule
\rowcolor{gray!15}
\textbf{Category} & \textbf{Metric} & \textbf{Linguistic Definition \& Rationale} \\
\midrule

\rowcolor{gray!5}
\textbf{Primary} 
& \textbf{Recovery Rate} 
& Exact match of canonical order $x$. Measures global structural integrity. \\

\midrule

\multirow{5}{*}{\textbf{Diagnostics}} 
& $S_{Rob}$ 
& Robustness score based on degradation under $S_n$ permutations (MDR/MDA). \\
\cmidrule(lr){2-3}
& \textbf{$S_{Acc}^{mean}$}
& Semantic accuracy via reference-meaning alignment using Cross-Encoders and BERTScore. \\
\cmidrule(lr){2-3}
& $S_{Logic}$ 
& NLI-based entailment check between model output $e$ and reference meaning. \\
\cmidrule(lr){2-3}
& $S_{Cons}$ 
& Stability across permutation variants, capturing invariance to internal shuffles. \\
\cmidrule(lr){2-3}
& $S_{Info}$ 
& Information density penalizing verbosity while rewarding high information-to-token ratio. \\

\bottomrule
\end{tabularx}
\vspace{-6pt}
\caption{\textbf{Evaluation Metrics on OrderProbe.} We prioritize Recovery Rate as the global structural indicator. Other diagnostic metrics characterize Robustness, Semantic Accuracy, Logical Validity, Structural Consistency, and Information Density.}
\label{app:metric_formulas}
\vspace{-0.22cm}
\end{table}

\section{Evaluation Framework}
\label{sec:eval_framework}

Beyond deterministic exact-match reconstruction, OrderProbe introduces a multi-dimensional diagnostic framework that separates structural reconstruction from meaning-oriented generation. 
A single aggregate score is often insufficient: models may fail recovery due to hallucinated retrieval, positional sensitivity, format mimicry, or verbose ``knowledge dumping,'' all of which can mask the underlying failure mechanism. 
We therefore design six complementary metrics(Table~\ref{app:metric_formulas}) computed over the full permutation space, enabling fine-grained failure-mode analysis.

\vspace{-6pt}
\paragraph{Evaluation Scope.}
For each canonical expression $x$, we evaluate model behavior under all 23 non-identity permutations $\mathcal{P}(x)$, yielding $3543 \times 23 = 81{,}489$ perturbed inputs. 
Each output is scored by exact reconstruction, semantic alignment, and logical entailment, and then aggregated across permutations to quantify sensitivity and invariance. 
This produces a six-dimensional diagnostic signature per model, rather than only a leaderboard rank.

\subsection{Primary Metric: Recovery Rate}
The Recovery Rate is the global indicator of structural integrity:
\vspace{-6pt}
\begin{equation}
\text{Recovery} = \frac{1}{N} \sum_{i=1}^N \mathbb{I}(\hat{x}_i = x_i).
\end{equation}
Unlike sentence restoration (ill-posed due to multiple valid reorderings), four-character expressions have a unique canonical form, enabling deterministic evaluation.

\subsection{Diagnostic Metrics (Beyond Recovery)}
We introduce five diagnostic dimensions to explain \textit{why} recovery fails.

\vspace{-6pt}
\paragraph{Semantic Accuracy ($S^{mean}_{Acc}$).}
We compare each model explanation with the validated reference meanings using a tiered hybrid score that combines cross-encoder relevance, multilingual embedding similarity, and lexical safeguards against entity-level hallucination. For each semantic component, we retain the best match over the valid references, so a paraphrase is not penalized for failing to match another valid wording (Appendix~\ref{app:sem_acc}).

\vspace{-6pt}
\paragraph{Logical Validity ($S_{Logic}$).}
To detect fluent but contradictory definitions, we compute multilingual-NLI entailment in both directions between the model explanation and each reference meaning, retain the higher directional score for each reference, and then take the best reference match. This measures compatibility with at least one validated meaning rather than strict equivalence (Appendix~\ref{app:logical}).
\vspace{-6pt}
\begin{equation}
S_{\text{Logic}} = \max_{r\in\mathcal{R}_x}\max\{P_{\text{NLI}}(e\Rightarrow r),P_{\text{NLI}}(r\Rightarrow e)\}.
\end{equation}

\paragraph{Structural Consistency ($S_{Cons}$).}
Structural consistency measures whether a model preserves similar semantic behavior across different permutations of the same canonical expression. For each expression $x$, we compute the semantic score of the model output under every perturbed input in $\mathcal{P}(x)$ and summarize the dispersion of these scores across permutations. We define:
\vspace{-6pt}
\begin{equation}
S_{\text{Cons}} = (1 - E_{\text{perf}})\cdot(1 - R_{\text{sens}}),
\end{equation}
where $E_{\text{perf}}$ captures the average deviation from the best-performing permutation and $R_{\text{sens}}$ captures the worst-case deviation. This metric penalizes both global instability and localized brittleness under internal reordering (Appendix~\ref{app:consistency}).

\begin{table*}[t!]
\centering
\footnotesize
\renewcommand{\arraystretch}{1.05}
\setlength{\tabcolsep}{8pt}

\setlength{\aboverulesep}{0pt}
\setlength{\belowrulesep}{0pt}

\begin{tabular}{ll ccccc c}
\toprule
\multirow{2}{*}{\textbf{Model Family}} & \multirow{2}{*}{\textbf{Setting}}
& \multicolumn{5}{c}{\textbf{Diagnostic Component Metrics}} & \textbf{Main Metric} \\
\cmidrule(lr){3-7} \cmidrule(lr){8-8}
& & $S_{Acc}^{mean}$ & $S_{Cons}$ & $S_{Logic}$ & $S_{Rob}$ & $S_{Info}$ & \textbf{Recovery. (\%)} \\
\midrule

\multicolumn{8}{l}{\cellcolor{gray!8}\textbf{Tier 1: High-Performance Reasoning Models}} \\

\multirow{2}{*}{Qwen-3-14B}
& Zero-shot & 0.493 & 0.591 & 0.672 & 0.621 & \textbf{0.579} & 31.062 \\
& \cellcolor{blue!5}+CoT
& \cellcolor{blue!5}0.429 & \cellcolor{blue!5}0.741 & \cellcolor{blue!5}0.811 & \cellcolor{blue!5}0.603 & \cellcolor{blue!5}0.536
& \cellcolor{blue!5}31.527 {\color{green!60!black} ($\uparrow$0.465)} \\

\multirow{2}{*}{Qwen-3-VLThink}
& Zero-shot & \textbf{0.520} & 0.539 & 0.712 & 0.645 & 0.520 & 30.928 \\
& \cellcolor{blue!5}+CoT
& \cellcolor{blue!5}0.471 & \cellcolor{blue!5}0.716 & \cellcolor{blue!5}0.888 & \cellcolor{blue!5}0.632 & \cellcolor{blue!5}0.533
& \cellcolor{blue!5}48.376 {\color{green!60!black} ($\uparrow$17.448)} \\

\multirow{2}{*}{Gemini-2.5-Flash}
& Zero-shot & 0.493 & 0.621 & 0.646 & 0.631 & 0.540 & 30.487 \\
& \cellcolor{blue!5}+CoT
& \cellcolor{blue!5}0.455 & \cellcolor{blue!5}0.738 & \cellcolor{blue!5}0.738 & \cellcolor{blue!5}\textbf{0.647} & \cellcolor{blue!5}0.246
& \cellcolor{blue!5}41.559 {\color{green!60!black} ($\uparrow$11.072)} \\

\midrule

\multicolumn{8}{l}{\cellcolor{gray!8}\textbf{Tier 2: Mid-Range General Models}} \\

\multirow{2}{*}{GPT-4o}
& Zero-shot & 0.463 & 0.663 & 0.591 & 0.616 & 0.489 & 23.254 \\
& \cellcolor{blue!5}+CoT
& \cellcolor{blue!5}0.431 & \cellcolor{blue!5}0.769 & \cellcolor{blue!5}0.774 & \cellcolor{blue!5}0.622 & \cellcolor{blue!5}0.400
& \cellcolor{blue!5}37.687 {\color{green!60!black} ($\uparrow$14.433)} \\

\multirow{2}{*}{GLM-4VThink}
& Zero-shot & 0.446 & 0.565 & 0.629 & 0.591 & 0.446 & 21.636 \\
& \cellcolor{blue!5}+CoT
& \cellcolor{blue!5}0.413 & \cellcolor{blue!5}0.785 & \cellcolor{blue!5}0.756 & \cellcolor{blue!5}0.594 & \cellcolor{blue!5}0.361
& \cellcolor{blue!5}30.079 {\color{green!60!black} ($\uparrow$8.443)} \\

\multirow{2}{*}{DeepSeek-V3.2}
& Zero-shot & 0.472 & 0.593 & 0.575 & 0.609 & 0.444 & 20.104 \\
& \cellcolor{blue!5}+CoT
& \cellcolor{blue!5}0.482 & \cellcolor{blue!5}0.727 & \cellcolor{blue!5}\textbf{0.948} & \cellcolor{blue!5}0.643 & \cellcolor{blue!5}\textbf{0.689}
& \cellcolor{blue!5}\textbf{62.238} {\color{green!60!black} ($\uparrow$42.134)} \\

\midrule

\multicolumn{8}{l}{\cellcolor{gray!8}\textbf{Tier 3: Efficient \& Baseline Models}} \\

\multirow{2}{*}{DeepSeek-R1}
& Zero-shot & 0.446 & 0.682 & 0.465 & 0.593 & 0.469 & 18.596 \\
& \cellcolor{blue!5}+CoT
& \cellcolor{blue!5}0.420 & \cellcolor{blue!5}0.790 & \cellcolor{blue!5}0.739 & \cellcolor{blue!5}0.612 & \cellcolor{blue!5}0.470
& \cellcolor{blue!5}27.729 {\color{green!60!black} ($\uparrow$9.133)} \\

\multirow{2}{*}{Claude-HaiKu-4.5}
& Zero-shot & 0.448 & 0.656 & 0.494 & 0.612 & 0.466 & 18.017 \\
& \cellcolor{blue!5}+CoT
& \cellcolor{blue!5}0.398 & \cellcolor{blue!5}0.714 & \cellcolor{blue!5}0.652 & \cellcolor{blue!5}0.551 & \cellcolor{blue!5}0.293
& \cellcolor{blue!5}21.646 {\color{green!60!black} ($\uparrow$3.629)} \\

\multirow{2}{*}{Gemma-3-4B}
& Zero-shot & 0.439 & 0.654 & 0.578 & 0.610 & 0.291 & 17.094 \\
& \cellcolor{blue!5}+CoT
& \cellcolor{blue!5}0.402 & \cellcolor{blue!5}0.783 & \cellcolor{blue!5}0.655 & \cellcolor{blue!5}0.600 & \cellcolor{blue!5}0.104
& \cellcolor{blue!5}18.027 {\color{green!60!black} ($\uparrow$0.933)} \\

\multirow{2}{*}{Llama-3-8B}
& Zero-shot & 0.396 & 0.706 & 0.399 & 0.590 & 0.304 & 12.167 \\
& \cellcolor{blue!5}+CoT
& \cellcolor{blue!5}0.385 & \cellcolor{blue!5}\textbf{0.815} & \cellcolor{blue!5}0.641 & \cellcolor{blue!5}0.586 & \cellcolor{blue!5}0.226
& \cellcolor{blue!5}24.428 {\color{green!60!black} ($\uparrow$12.261)} \\

\multirow{2}{*}{Llama-2-7B}
& Zero-shot & 0.411 & 0.731 & 0.502 & 0.588 & 0.182 & 12.576 \\
& \cellcolor{blue!5}+CoT
& \cellcolor{blue!5}0.412 & \cellcolor{blue!5}0.744 & \cellcolor{blue!5}0.469 & \cellcolor{blue!5}0.602 & \cellcolor{blue!5}0.258
& \cellcolor{blue!5}18.747 {\color{green!60!black} ($\uparrow$6.171)} \\

\multirow{2}{*}{Qwen-3-8B}
& Zero-shot & 0.446 & 0.617 & 0.487 & 0.604 & 0.403 & 15.786 \\
& \cellcolor{blue!5}+CoT
& \cellcolor{blue!5}0.392 & \cellcolor{blue!5}0.715 & \cellcolor{blue!5}0.492 & \cellcolor{blue!5}0.580 & \cellcolor{blue!5}0.269
& \cellcolor{blue!5}12.594 {\color{red} ($\downarrow$3.192)} \\

\bottomrule
\end{tabular}
\vspace{-0.25cm}
\caption{Comparative Assessment of Zero Shot and Chain of Thought Prompting across 12 LLMs on OrderProbe. This table reports performance metrics encompassing semantic accuracy, consistency and structural recovery.}
\label{tab:comprehensive_performance_final}
\vspace{-0.6cm}
\end{table*}

\vspace{-6pt}
\enlargethispage{-\baselineskip}
\paragraph{Robustness ($S_{Rob}$).}
We quantify robustness as degradation relative to the canonical baseline and decompose it into sequential robustness (permutation sensitivity) and structural robustness (variance across syntactic categories), combined via harmonic mean (Appendix~\ref{app:robustness}).

\vspace{-8pt}
\paragraph{Information Density ($S_{Info}$).}
To penalize verbosity-based metric inflation, we introduce an information density score using a brevity penalty that rewards informative explanations per token (Appendix~\ref{app:density}). We additionally test the robustness of the diagnostic pipeline under controlled substitutions of major components, including the cross-encoder, embedding similarity, NLI model, aggregation strategy, and weight scheme. Model rankings remain stable across variants, with Spearman correlations ranging from 0.842 to 0.931 relative to the original configuration, suggesting that the main conclusions are not driven by a particular metric choice (Appendix~\ref{app:metric_robustness}).

\section{Experiments}

\begin{figure*}[t!]
    \centering
    \includegraphics[width=1\linewidth]{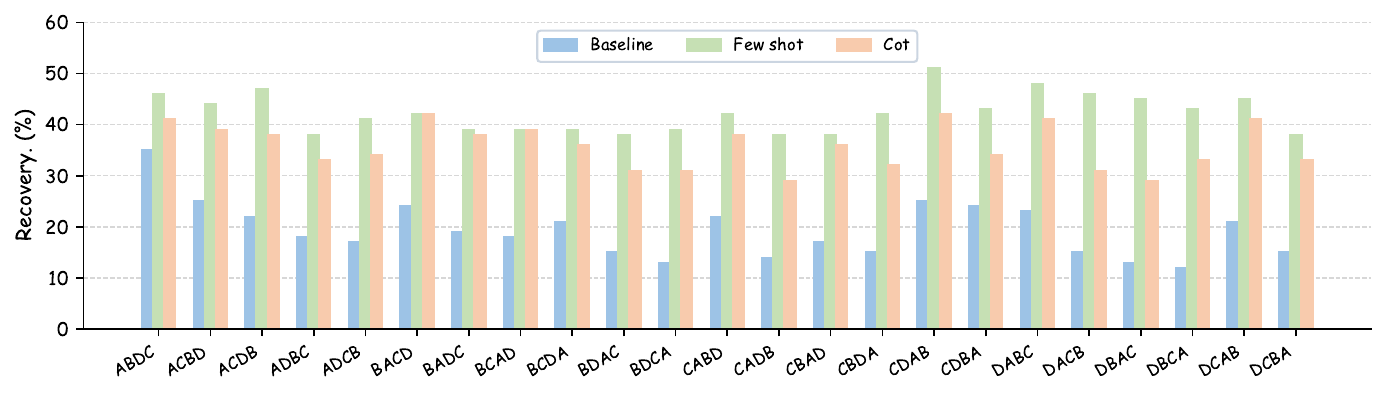}
    \vspace{-0.6cm}
    \caption{\textbf{Performance Comparison across 23 Permutation Patterns.} The chart illustrates recovery rates under baseline, few shot, and chain of thought settings, revealing the impact of anchor displacement on model stability.}
    \label{figrepro}
    \vspace{-0.4cm}
\end{figure*}

\subsection{Experimental Setup}
\label{sec:experimental_setup}
We evaluate all models on the full OrderProbe benchmark containing $3{,}543$ canonical four-character expressions across four script settings (ZH-CN, ZH-TW, JA, KO).
Given each scrambled input, models are required to output \textit{(i)} the recovered canonical expression (a single four-character sequence) and \textit{(ii)} a one-line semantic explanation in the corresponding language.
All outputs are evaluated by exact-match reconstruction and our diagnostic metrics described in Section~\ref{sec:eval_framework}.
We additionally evaluate a reconstruction-only output schema in which models provide only the recovered expression; results are reported in Appendix~\ref{app:reconstruction_only}.

\vspace{-6pt}
\paragraph{Models.}
We evaluate twelve widely used LLMs spanning open- and closed-source families.
Open-source models include instruction-tuned checkpoints from Qwen \citep{qwen2025qwen25technicalreport}, DeepSeek \citep{deepseekai2025deepseekv3technicalreport}, Llama \citep{grattafiori2024llama3herdmodels}, and Gemma, run locally using official weights.
Closed-source models include GPT \citep{openai2024gpt4technicalreport}, Gemini, Claude \citep{anthropic_claude}, and GLM \citep{5team2025glm45agenticreasoningcoding}, accessed via official APIs with provider-default settings during Oct 2025--Jan 2026.
All checkpoint and deployment details are provided in Appendix~\ref{app:deployment}.

\vspace{-6pt}
\paragraph{Prompting Strategy and Output Constraints.}
We use a unified two-message prompting format (system + user) with strict schema enforcement to minimize format drift.
We adopt language-matched prompting: each script setting is evaluated using prompts written in the same language as the input, and models are instructed to produce both the recovered expression and explanation in that language.
We compare three settings: \textbf{(i)} Zero-shot (one-line output), \textbf{(ii)} Chain-of-Thought (CoT; exactly two lines: reasoning + final answer), and \textbf{(iii)} three-shot in-context learning (few-shot; two-line output following exemplars).
All prompt templates are provided in Appendix~\ref{sec:appendix_prompts}.
English functional readings of the multilingual templates are provided in Appendix~\ref{app:prompt_translations}.

\begin{table}[t]
\centering
\small
\renewcommand{\arraystretch}{1.3}
\setlength{\tabcolsep}{4pt}

\setlength{\aboverulesep}{0pt}
\setlength{\belowrulesep}{0pt}

\resizebox{\linewidth}{!}{
\begin{tabular}{@{}l | ccccc | >{\bfseries\color{HSSLightBlue}}c@{}}
\toprule

\rowcolor{HSSGray}
& \multicolumn{5}{c|}{\textbf{Component Metrics}} 
& \multicolumn{1}{c}{\textbf{Main Metric}} \\

\rowcolor{HSSGray}
\multirow{-2}{*}{\textbf{Language}} 
& $S_{Acc}^{mean}$ & $S_{Cons}$ & $S_{Logic}$ & $S_{Rob}$ & $S_{Info}$ 
& \multicolumn{1}{c}{\textbf{Recov. (\%)}} \\
\midrule

Simplified Chinese   & \textbf{0.4817} & 0.6445 & \textbf{0.6854} & \textbf{0.6193} & 0.4631 & 25.20 \\
Traditional Chinese  & 0.4715 & 0.6184 & 0.6663 & 0.5998 & 0.4040 & 20.32 \\
Japanese             & 0.4602 & 0.5897 & 0.5278 & 0.6048 & \textbf{0.5388} & 19.66 \\
\midrule

\rowcolor{blue!6}
Korean               & 0.3990 & \textbf{0.7625} & 0.3014 & 0.5857 & 0.3906 & 5.72 \\

\bottomrule
\end{tabular}
}

\caption{\textbf{Comparison across script settings.} Logographic scripts (Chinese/Japanese) yield substantially higher reconstruction accuracy than Korean Hangul, which serves as a phonogrammatic control condition. Component metrics are reported for diagnostic analysis alongside the primary Recovery score.}
\label{tab:lang_comparison}
\vspace{-0.22 cm}
\end{table}

\subsection{Overall Results and Analysis}
\label{sec:overall_results}

Table~\ref{tab:comprehensive_performance_final} summarizes the benchmark results across twelve models under both Zero-shot and CoT settings.
Across model families, we observe a consistent gap between meaning-oriented generation and exact structural recovery: models can often generate fluent, semantically plausible explanations via lexical retrieval from partially preserved cues, while still failing to reconstruct the canonical order under permutation noise.

\vspace{-6pt}
\paragraph{Reconstruction remains challenging.}
Exact reconstruction remains difficult even for strong systems, with zero-shot recovery frequently below 35\%. This suggests that structural recovery is not an automatic byproduct of semantic competence.
Among non-exact zero-shot outputs, related but non-canonical expressions account for 12.80\%, while literal recombinations account for 31.50\% (Appendix~\ref{app:error_categories}); these near-misses do not satisfy the exact canonical-order criterion.
Human participants recover 46.5--88.6\% of the same scrambled inputs across scripts, whereas the canonical-input control reaches 100\% Recovery by construction; its diagnostic scores evaluate the associated model explanation separately (Appendices~\ref{tab:human_reconstruction_baseline} and~\ref{sec:appendix_reference_performance}).

\vspace{-6pt}
\paragraph{CoT helps, but gains vary widely.}
CoT prompting often improves recovery, suggesting that explicit intermediate reasoning can encourage constraint checking before generation. However, the gains are strongly model-dependent and can reverse under unstable decoding.

\vspace{-6pt}
\paragraph{Consistency is not correctness.}
High structural consistency can coincide with low recovery, as models may produce stable but generic explanations across permutations. This reinforces the need to interpret diagnostic metrics jointly.

\vspace{-6pt}
\paragraph{Summary.}
Overall, OrderProbe shows that models can retrieve meanings or provide plausible definitions while still failing exact canonical reconstruction.

\subsection{Permutation Sensitivity and Prompting Effects}
\label{sec:perm_sensitivity}

Figure~\ref{figrepro} compares recovery trends across all 23 permutation configurations under baseline, CoT, and few-shot settings.
We observe strong dependence on anchor displacement: permutations that preserve semantically informative characters near their canonical neighborhoods yield higher recovery, while dispersed anchors significantly degrade performance.

Few-shot prompting yields the most uniform improvements across permutations, suggesting that explicit exemplars constrain decoding more effectively than generic reasoning traces.
In contrast, CoT gains are less stable and may amplify format drift in weaker models.

\subsection{Impact of Script Typology}
\label{sec:script_typology}

Table~\ref{tab:lang_comparison} compares performance across the four script settings.
We observe that reconstruction difficulty is strongly shaped by script typology.

\vspace{-6pt}
\paragraph{Logographic scripts provide stronger local anchors.}
Chinese and Japanese exhibit substantially higher recovery than Korean.
Logographic characters preserve standalone semantic cues under scrambling, providing local anchors that facilitate global reassembly.

\vspace{-6pt}
\paragraph{Hangul is a phonogrammatic control.}
Korean Hangul behaves differently: syllable-block scrambling preserves syllable composition but weakens standalone semantic anchors.
As a result, recovery collapses sharply, suggesting that reconstruction depends not only on positional decoding but also on interpretable local anchors.

\vspace{-6pt}
\paragraph{Consistency inflation under extreme ambiguity.}
Notably, Korean yields high consistency despite low recovery.
Under severe uncertainty, models often default to generic, stable explanations across permutations.
This yields artificially high invariance but does not reflect true reconstruction ability, reinforcing the need to interpret diagnostic metrics jointly.

\begin{figure}[t]
    \centering
    \includegraphics[width=1\linewidth]{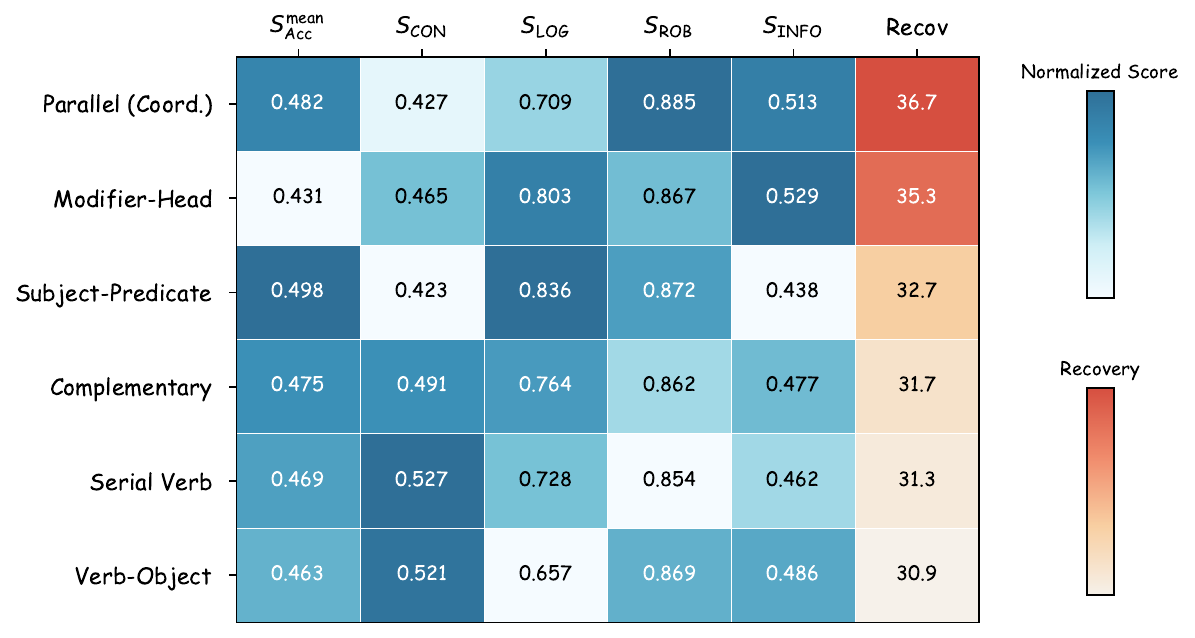}
    \caption{\textbf{Performance by syntactic category.} Parallel structures are easiest due to redundancy and symmetry cues, while asymmetric dependency patterns remain harder for reconstruction. Metrics are reported to diagnose whether failures arise from semantic gaps, instability, or structural sensitivity.}
    \label{fig:syntactic}
\end{figure}

\subsection{Analysis of Syntactic Patterns}
\label{sec:structure_patterns}

Figure~\ref{fig:syntactic} reports results across six syntactic categories, revealing systematic structural variation beyond surface shuffling.

\vspace{-6pt}
\paragraph{Parallel structures are easiest.}
Parallel and coordinate patterns are generally easier due to redundancy and symmetric templates.
These cues effectively reduce the search space and provide ordering heuristics.

\vspace{-6pt}
\paragraph{Meaning does not imply correct ordering.}
For subject--predicate patterns, models often recover the underlying proposition but fail to linearize characters into the precise canonical order, confirming that semantic access does not guarantee successful constraint-based sequence reconstruction.

\vspace{-6pt}
\paragraph{Asymmetric dependencies remain difficult.}
Verb--object patterns are harder because they rely on directional dependencies that are disrupted by permutation noise, leading to semantically plausible but structurally incorrect guesses.

\begin{figure}[t!]
    \centering
    \includegraphics[width=1\linewidth]{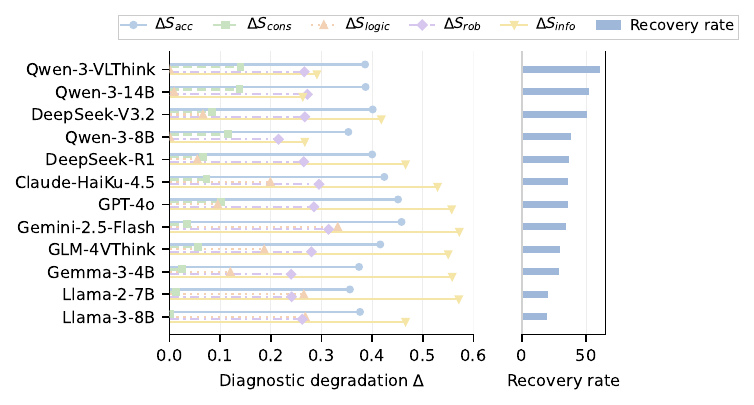}
    \caption{\textbf{Degradation under perturbed 3-shot inputs.}
Left: per-metric drops $\Delta S_{acc}$, $\Delta S_{cons}$, $\Delta S_{logic}$, $\Delta S_{rob}$, $\Delta S_{info}$, computed as \emph{unperturbed} minus \emph{perturbed 3-shot} under matched inference settings.
Right: recovery rate for perturbed 3-shot inputs.
Models are ordered by recovery rate (higher is more robust).}

    \label{fig:fewshot_degradation}
\end{figure}

\subsection{Analysis of In-Context Learning}
\label{sec:few_shot}

Figure~\ref{fig:fewshot_degradation} reports results under three-shot prompting.

\vspace{-6pt}
\paragraph{Perturbations reveal structured degradation.}
Across models, diagnostic drops concentrate on particular components rather than moving in lockstep, suggesting that perturbations disrupt specific capabilities rather than inducing uniform instability.

\vspace{-6pt}
\paragraph{High recovery does not imply preserved diagnostics.}
Some models retain relatively strong recovery while showing clear drops on particular diagnostics, whereas low-recovery models typically decline across multiple dimensions.

\subsection{Case Study: Literal Interpretation under Structural Perturbation}
\label{sec:case_study}

\begin{figure}[t]
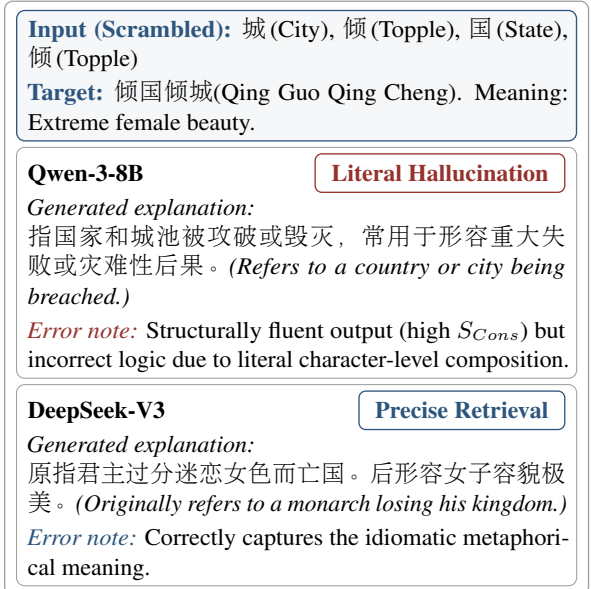

\centering
\small
\begin{CJK*}{UTF8}{gbsn}

\begin{tcolorbox}[
  compactbox,
  width=\linewidth,
  colback=white,
  colframe=LineGray,
  boxrule=0.5pt,
  arc=1mm,
  left=4pt,right=4pt,top=3pt,bottom=3pt
]

\begin{tcolorbox}[
  compactbox,
  colback=SoftGray,
  colframe=MainBlue,
  boxrule=0.5pt,
  arc=1mm,
  left=4pt,right=4pt,top=2pt,bottom=2pt
]
\setstretch{1.12}

\textbf{\textcolor{MainBlue}{Input (Scrambled):}}
城\,(City), 倾\,(Topple), 国\,(State), 倾\,(Topple)

\vspace{1.5pt}
\textbf{\textcolor{MainBlue}{Target:}}
倾国倾城 (Qing Guo Qing Cheng). Meaning: Extreme female beauty.
\end{tcolorbox}

\vspace{2pt} 

\begin{tcolorbox}[
  compactbox,
  colback=white,
  colframe=LineGray,
  boxrule=0.4pt,
  arc=1mm,
  left=4pt,right=4pt,top=2pt,bottom=2pt
]
\setstretch{1.12}

\textbf{Qwen-3-8B}\hfill \labeltag{WarnRed}{Literal Interpretation}

\vspace{1.5pt}
\textit{Generated explanation:}

指国家和城池被攻破或毁灭，常用于形容重大失败或灾难性后果。  
{\footnotesize\textit{(Refers to a country or city being breached.)}}

\vspace{1.5pt}
\textit{\textcolor{WarnRed}{Error note:}}
The model produces a literal compositional interpretation based on
individual characters rather than retrieving the conventional idiomatic meaning.
Although the explanation is locally plausible, it does not correspond to the
canonical lexicalized expression.

\end{tcolorbox}

\vspace{2pt} 

\begin{tcolorbox}[
  compactbox,
  colback=white,
  colframe=LineGray,
  boxrule=0.4pt,
  arc=1mm,
  left=4pt,right=4pt,top=2pt,bottom=2pt
]
\setstretch{1.12}

\textbf{DeepSeek-V3}\hfill \labeltag{MainBlue}{Precise Retrieval}

\vspace{1.5pt}
\textit{Generated explanation:}

原指君主过分迷恋女色而亡国。后形容女子容貌极美。  
{\footnotesize\textit{(Originally refers to a monarch losing his kingdom.)}}

\vspace{1.5pt}
\textit{\textcolor{MainBlue}{Error note:}}
Correctly captures the idiomatic metaphorical meaning.
\end{tcolorbox}

\end{tcolorbox}

\end{CJK*}
\vspace{1.5pt}
\caption{\textbf{Case Study on Literal Interpretation.}
Qwen-3-8B produces a fluent but non-canonical literal interpretation, while DeepSeek-V3 correctly retrieves the idiomatic metaphor.}
\vspace{-10pt}
\label{fig:case_study}
\end{figure}

Figure~\ref{fig:case_study} shows a representative case on \textit{Qing Guo Qing Cheng}. Some models retrieve the idiomatic metaphor correctly, while others produce a plausible but non-canonical literal interpretation. This example illustrates why exact reconstruction and diagnostic metrics are both necessary.

\subsection{Memorization and Frequency Analysis}
\label{sec:freq_analysis}

A natural concern is whether reconstruction mainly reflects memorization of frequent expressions. We therefore conduct a pilot frequency-stratified analysis (Appendix~\ref{app:freq_pilot}). Although recovery improves with idiom frequency, failures remain common even in high-frequency bins, suggesting that lexical familiarity alone does not guarantee successful reconstruction.

Root-grouped regression further shows that frequency contributes to Recovery, while structural predictors and script/model controls provide additional explanatory power (Appendix~\ref{app:freq_incremental}).

We analyze the role of lexical familiarity using a frequency-stratified experiment under progressively disrupted semantic signals. The frequency advantage diminishes as more signals are removed (Table~\ref{tab:freq_disruption}).

To further examine whether reconstruction performance mainly reflects memorized idiom retrieval, we construct a synthetic control set consisting of non-attested four-character expressions. For each synthetic item, a canonical order is defined and the same permutation protocol and evaluation procedure are applied.

Models continue to exhibit low reconstruction accuracy on synthetic expressions, although performance is slightly lower than on idiomatic items. This result suggests that lexicalized idiom knowledge can provide partial cues, but the difficulty observed in OrderProbe cannot be reduced to memorized idiom lookup alone. Detailed construction procedures and quantitative results are provided in Appendix~\ref{app:synthetic_control} and Table~\ref{tab:synthetic_control}.

Prior work on typoglycemia, the recognition of words or expressions despite internal character transpositions when surface cues remain available, shows that scrambled text can often preserve semantic accessibility even when exact internal order is disrupted \citep{yu2024mindscrambleunveilinglarge}. This mechanism is consistent with our observation that models frequently generate meaning-aligned explanations while still failing exact canonical recovery. At the same time, typoglycemia-style robustness is often driven by lexical activation from partially preserved surface form \citep{wang2025wordformmattersllms}, whereas our synthetic-control results show that reconstruction remains difficult even when lexicalized idiom knowledge is reduced. This suggests that OrderProbe probes a stricter setting in which semantic access may survive scrambling, but exact canonical recovery still requires constraint-based reconstruction beyond lexical activation alone.

\section{Related Work}
\label{sec:related_work}

Prior robustness benchmarks have examined lexical noise, adversarial perturbations, instruction manipulation, and reasoning under altered inputs. Early resources such as TextFlint \citep{wang-etal-2021-textflint} and CheckList \citep{ribeiro-etal-2020-beyond} mainly focus on surface-level variation, while later work extends robustness evaluation to instruction following \citep{li-etal-2024-evaluating-instruction} and logical reasoning \citep{lin-etal-2025-assessing}. Related studies on typoglycemia and scrambled text ask whether models can still recognize or understand perturbed inputs. In contrast, OrderProbe does not test robustness to generic noise, but exact reconstruction of a unique canonical form from internally permuted components \cite{yang2026trilens,su2026revisiting,li2026personalize,li2025miv,li2025chemvlm,li2026faithful,li2025ddtime,li2026rethinking,wu2026roboalign,yu2026composition,yu2026cordon,yu2026detecting,yu2026fides}.

Research on idiom and figurative language processing has largely treated idioms as static lexical units. Benchmarks such as ChID \citep{zheng-etal-2019-chid}, MMFLD \citep{lai-etal-2023-multilingual}, Chengyu-Bench \citep{fu2025chengyubenchbenchmarkinglargelanguage}, and Idiom-Trans \citep{donthi-etal-2025-improving} mainly evaluate cloze prediction, interpretation, translation, or usage under canonical inputs. Classical-language benchmarks such as WenMind \citep{NEURIPS2024_5c1019b5} similarly emphasize semantic understanding rather than recovery under internal perturbation. By contrast, OrderProbe focuses on whether models can recover the conventional internal order of lexicalized four-character expressions, enabling deterministic exact-match evaluation \cite{ye-etal-2026-tableqa, ye2026rethinkingstepwisemodelrouting,ye2026rubricguidedprocessrewardstepwise, guo-etal-2026-rethinking-table,guo-etal-2025-sqlforge,luo2026biasig,luo2024faintbench,luo2024versusdebias,li2026prefix,yu2026robosemanticbench,yu2026twinbrainvla,DBLP:journals/corr/abs-2604-03647,zhang2026self,zheng2026trajectory,zheng2026maigo,zheng2026pilot}. 

Our findings are also related to work on positional bias and structural limitations in Transformer-based models. \citet{sinha-etal-2021-unnatural} show that models can be unexpectedly insensitive to word order in some settings, and \citet{berglund2024the} document failures to generalize relational structure under reversal \cite{10.1145/3774904.3792075,kang2026hssbench,kang2026multimodal,kang2026quanteval,feng2025seeing,wang2025mucar,liu2026reasonact,gao2026laobench,shi2026spader,chen2026diagdiagnosticiterativealignment,chen2026learning}. OrderProbe complements these findings with a controlled benchmark in Chinese, Japanese, and Korean, showing that even when local semantic cues remain partially available, exact canonical reconstruction under internal permutation remains challenging for current LLMs.

\section{Conclusion}

We introduce \textbf{OrderProbe}, a deterministic benchmark for structural reconstruction in Chinese, Japanese, and Korean four-character expressions. Experiments on twelve LLMs show that exact recovery remains difficult, with zero-shot performance often below 35\%, highlighting a gap between semantic access and exact structural reconstruction. We also find strong effects of script typology: logographic scripts offer stronger local anchors, whereas Hangul provides a phonogrammatic control with weaker local anchoring. OrderProbe is intended to complement broader evaluations of order sensitivity, and we release the benchmark and diagnostic framework to support future research on structural robustness in LLMs.

\section*{Limitations}
OrderProbe evaluates structural reconstruction in a controlled lexical setting, but it does not cover the full range of linguistic structures. Our analysis is limited to six syntactic categories and to fixed four-character expressions in three languages. We also do not study alphabetic or morphologically rich systems in detail.

In addition, some expressions are likely to appear in pretraining corpora, so reconstruction may partly benefit from memorized lexical knowledge. Although our frequency analysis suggests that memorization alone does not explain the results, broader tests on novel, synthetic, or non-idiomatic controls remain for future work.

\section*{Ethics Statement}

The OrderProbe dataset integrates linguistic resources from the public domain under fair use principles for academic research. We confirm that the collected expressions are free from copyright restrictions and strictly adhere to data usage policies. All participants provided informed consent and followed a rigorous protocol to exclude offensive or discriminatory content.

Our data construction pipeline utilizes LLMs for semantic augmentation in full compliance with provider terms of service. We declare that all generated outputs underwent human verification to ensure factual accuracy and safety. This work serves as a diagnostic benchmark for enhancing model robustness. We foresee no direct negative societal impacts or misuse risks associated with the release of this dataset.



\bibliography{custom}

\appendix

\section{Formal Definitions and Rationale of Evaluation Metrics}
\label{sec:appendix_metrics}  

This appendix details the mathematical formulation and design rationale for the evaluation metrics employed in OrderProbe. 
To avoid the ambiguity often introduced by aggregated leaderboards, we adopt a \textbf{multi-dimensional diagnostic framework}. 
Each metric is designed to capture a specific failure pattern—ranging from structural rigidity to semantic hallucination—thereby providing a granular dissection of model capabilities.
These metrics are reported independently in the main paper and are not aggregated into a single score.

\subsection{Primary Metric: Recovery Rate}
\label{app:rec_acc}

\paragraph{Rationale}
The ability to recover the canonical form serves as a direct indicator of structural reconstruction ability. 
Unlike semantic comprehension, which can be inferred probabilistically from keywords, structural reconstruction requires resolving specific positional dependencies. 
Strict matching is necessary to distinguish exact canonical reconstruction from partial meaning recovery.

\paragraph{Formulation}
Let $x$ denote a canonical four-character idiom and $x'=\mathcal{T}(x,\pi)$ its internally reordered variant. We write a model output as $o=(\hat{x},e)$, where $\hat{x}$ is the recovered expression and $e$ is the model explanation.
Reconstruction is marked as correct only if $\hat{x}$ exactly matches the canonical idiom $x$.
Partial matches, paraphrases, or explanations without exact reconstruction are not counted as correct.

The Recovery Rate is defined as:
\begin{equation}
\text{Recovery}= \frac{1}{N} \sum_{i=1}^{N} \mathbb{I}(\hat{x}_i = x_i),
\end{equation}
In this context, $\mathbb{I}(\cdot)$ represents the indicator function.

\subsection{Semantic Accuracy}
\label{app:sem_acc}

\paragraph{Rationale and Component Selection}
Assessing idiom interpretation requires capturing both deep metaphorical meaning and precise entity details.
Single-dimensional metrics cannot capture this duality.
Vector-based similarity measures often miss specific entity errors, while n-gram overlap fails to recognize valid metaphorical paraphrases.
To address this, we construct a tiered hybrid metric, $S_{\text{Acc}}$, that integrates signals from both interaction-based and lexical levels.

\paragraph{Weighting Justification}
We assign weights to reflect the functional hierarchy of each verification method.
We treat the cross-encoder as the primary evaluator and assign it a weight of $0.5$.
Unlike bi-encoder architectures, which compress sentences into independent vectors, cross-encoders perform full self-attention over the concatenated input.
This enables them to detect subtle semantic entailments that are essential for precise interpretation.
We assign a moderate weight of $0.3$ to an ensemble of embedding-based measures, including BERTScore, STS, and cosine similarity, which serves as a robust semantic baseline.
Relying on a single embedding model may introduce architectural bias.
Integrating distinct representation approaches reduces variance and provides a more general measure of semantic proximity.
We allocate the remaining weight of $0.2$ to the $F_{\beta}$ score as a lexical safeguard.
Models can exhibit high similarity scores while hallucinating or mismatching specific entities.
This lexical component therefore penalizes factual discrepancies.

\paragraph{Formulation}
For an explanation $e$ and a reference meaning $r$, the composite Semantic Accuracy score is defined as:
\begin{equation}
\begin{split}
S_{\text{Acc}} = w_1 \cdot S'_{\text{ce}}
              & + w_2 \cdot \left(\frac{S'_{\text{bert}} + S'_{\text{sts}} + S'_{\text{cos}}}{3}\right) \\
              & + w_3 \cdot S_{f\beta}.
\end{split}
\end{equation}
$S'_{\text{ce}}$ denotes the normalized score produced by the cross-encoder.
The term in parentheses is the averaged embedding similarity score.
$S_{f\beta}$ denotes the lexical $F_{\beta}$ score.
We set the weights to $w_1 = 0.5$, $w_2 = 0.3$, and $w_3 = 0.2$ based on the hierarchy described above.
When an item has multiple validated reference meanings $\mathcal{R}_x$, each component score is computed against every $r\in\mathcal{R}_x$ and its best match is retained before combining components:
\begin{equation}
\begin{split}
S_{\text{Acc}}(e,\mathcal{R}_x) ={}& w_1\max_r S'_{\text{ce}}(e,r) \\
&+ \frac{w_2}{3}\max_r S'_{\text{bert}}(e,r) \\
&+ \frac{w_2}{3}\max_r S'_{\text{sts}}(e,r) \\
&+ \frac{w_2}{3}\max_r S'_{\text{cos}}(e,r) \\
&+ w_3\max_r S_{f\beta}(e,r), \\
&\hspace{2.3cm} r\in\mathcal{R}_x.
\end{split}
\end{equation}
Different components may therefore match different valid paraphrases; this avoids penalizing an explanation for not reproducing one particular reference wording.

\subsection{Structural Consistency}
\label{app:consistency}

\paragraph{Rationale}
Structural consistency evaluates whether a model preserves similar semantic behavior across different permutations of the same canonical expression. Since our evaluation uses deterministic decoding, each permuted input yields a single model output. We therefore measure consistency across the permutation set rather than across repeated generations of the same input.

\paragraph{Formulation}
Let $\mathcal{P}(x)$ denote the complete set of non-identity permutations for a canonical expression $x$. For each permutation $p \in \mathcal{P}(x)$, the model produces one output explanation, whose semantic accuracy score is denoted by $S_{\text{Acc}}(p)$.

We first define the best semantic performance observed across the permutation set:
\begin{equation}
S_{\max}(x) = \max_{p \in \mathcal{P}(x)} S_{\text{Acc}}(p).
\end{equation}

We then define the average deviation from this best-performing permutation as:
\begin{equation}
E_{\text{perf}}(x) = \frac{1}{|\mathcal{P}(x)|} \sum_{p \in \mathcal{P}(x)} \left(S_{\max}(x) - S_{\text{Acc}}(p)\right).
\end{equation}

We define the worst-case deviation as:
\begin{equation}
R_{\text{sens}}(x) = \max_{p \in \mathcal{P}(x)} \left(S_{\max}(x) - S_{\text{Acc}}(p)\right).
\end{equation}

Finally, we compute structural consistency as:
\begin{equation}
S_{\text{Cons}}(x) = (1 - E_{\text{perf}}(x)) \times (1 - R_{\text{sens}}(x)).
\end{equation}

This formulation assigns a high score to models whose semantic behavior remains stable across different internal reorderings of the same expression.

\subsection{Degradation Under Internal Reordering}
\label{app:degradation}

\paragraph{Rationale and Normalization Logic}
Isolating the impact of structural noise requires a performance baseline for each specific idiom.
We use the score obtained from the canonical non-reordered input, $S_{\text{orig}}$, as the reference capability.
A low score on a scrambled input indicates structural fragility only if the model demonstrates competence on the canonical form.
Direct comparison without normalization fails to distinguish between structural sensitivity and general knowledge deficits.
Consequently, we quantify degradation as a relative loss ratio.
MDR captures the expected utility loss under random structural perturbations.
MDA captures a lower bound on performance reliability, reflecting the worst-case structural collapse.
The maximization term in MDA extracts the worst-case signal across the permutation group to maintain a consistent definition.

\paragraph{Formulation}
We formalize the degradation metrics through the following equations.
$\mathcal{P}$ denotes the set of permutations applied to the idiom.
$S_{\text{orig}}$ represents the semantic accuracy score of the model on the canonical input.
$S_{\text{pert}}^{p}$ denotes the score under permutation $p$.
We define the Mean Degradation Rate (MDR) as the average relative performance drop across all permutations.
\begin{equation}
\text{MDR} = \frac{1}{|\mathcal{P}|} \sum_{p \in \mathcal{P}}
\frac{S_{\text{orig}} - S_{\text{pert}}^{p}}{S_{\text{orig}}}
\end{equation}
We define the Maximum Degradation Amount (MDA) as the maximum relative drop observed across the permutation group.
\begin{equation}
\text{MDA} = \max_{p \in \mathcal{P}}
\left(
\frac{S_{\text{orig}} - S_{\text{pert}}^{p}}{S_{\text{orig}}}
\right)
\end{equation}
These values quantify the average and worst-case loss of capability attributable to internal reordering.

\subsection{Robustness Metrics}
\label{app:robustness}

\paragraph{Rationale and Dual Dimensionality}
We assess robustness along two perturbation dimensions: Sequential Robustness and Structural Robustness.
Sequential Robustness measures the resilience of the model against internal ordering permutations.
Structural Robustness evaluates the stability of performance across distinct syntactic categories.
We select the mean semantic accuracy $S_{\text{Acc}}^{\text{mean}}$ as the basis for these metrics.
While maximum scores reflect peak recovery capability, average scores capture stability under stochastic degradation and thus provide a stricter and more sensitive indicator of robustness.
The final metric integrates these dimensions using a harmonic mean, ensuring that high robustness requires resilience to both combinatorial scrambling and syntactic variation.

\paragraph{Sequential Robustness Formulation}
We quantify Sequential Robustness $S_{\text{seq}}$ by combining the mean and maximum degradation rates defined in Appendix~\ref{app:degradation}.
We define $S_{\text{seq}}$ as the complement of a weighted sum of these degradations.
We assign equal weights $\alpha=0.5$ and $\beta=0.5$ to balance average stability and limit-state resilience.
\begin{equation}
S_{\text{seq}} = 1 - (\alpha \cdot \text{MDR} + \beta \cdot \text{MDA})
\end{equation}

\paragraph{Structural Robustness Formulation}
We quantify Structural Robustness $S_{\text{struct}}$ by measuring performance variance across syntactic categories.
Let $T$ denote the set of six syntactic structure types indexed by $k$.
For the $k$-th structure, we compute the intra-group mean accuracy $\mu_k$ over the idiom set $D_k$.
\begin{equation}
\mu_k = \frac{1}{|D_k|} \sum_{x \in D_k} S_{\text{Acc}}^{\text{mean}}(x)
\end{equation}
We calculate the standard deviation $\sigma$ of these group means to capture the sensitivity of the model to syntactic form.
We define $S_{\text{struct}}$ as the complement of the normalized deviation, ensuring that consistent performance across diverse structures yields a high score.
\begin{equation}
S_{\text{struct}} = 1 - \text{Normalize}(\sigma(\mu_1, \mu_2, ..., \mu_6))
\end{equation}

\paragraph{Composite Robustness Score}
We compute the final Robustness Score $S_{\text{Rob}}$ as the harmonic mean of the sequential and structural components.
This aggregation penalizes imbalance, requiring the model to demonstrate robustness in both internal ordering and syntactic generalization.
\begin{equation}
S_{\text{Rob}} = \frac{2 \cdot S_{\text{seq}} \cdot S_{\text{struct}}}{S_{\text{seq}} + S_{\text{struct}}}
\end{equation}

\subsection{Logical Validity}
\label{app:logical}

\paragraph{Rationale and Entailment Check}
High lexical similarity does not guarantee factual correctness.
Models may generate explanations that share keywords with the target while conveying contradictory meanings through negation or causal inversion.
Lexical metrics fail to detect these logical errors.
To address this, we employ a Natural Language Inference (NLI) model as a validity filter.
This component evaluates whether the generated explanation is logically compatible with a validated reference meaning, ensuring that high scores reflect true semantic understanding rather than superficial keyword matching.

\paragraph{Formulation}
We quantify logical validity $S_{\text{Logic}}$ as the probability of entailment predicted by a pretrained NLI model.
Given a generated explanation $e$ and a reference meaning $r$, we compute entailment in both directions:
\begin{equation}
S_{\text{Logic}}(e,r)=\max\{P_{\text{NLI}}(e\Rightarrow r),P_{\text{NLI}}(r\Rightarrow e)\}.
\end{equation}
The item-level score is $\max_{r\in\mathcal{R}_x}S_{\text{Logic}}(e,r)$, using the same best-reference convention as Semantic Accuracy. We use the maximum directional entailment because the diagnostic asks whether the explanation is compatible with at least one validated meaning; it is not intended as a strict equivalence test. Sensitivity to the unidirectional and bidirectional-mean alternatives is reported in Appendix~\ref{app:metric_robustness}.

\subsection{Information Density}
\label{app:density}

\paragraph{Rationale and Verbosity Penalty}
Models frequently employ knowledge-dumping strategies, generating excessively verbose explanations to maximize keyword coverage.
This behavior inflates recall-based metrics without reflecting precise understanding.
To counteract this, we introduce an Information Density metric that rewards high information content per token.
We impose a brevity penalty (BP) on outputs that exceed the reference length without proportional semantic gain.
This mechanism encourages the model to prioritize concise and accurate reasoning over redundant generation.

\paragraph{Formulation}
We compute Information Density $S_{\text{Info}}$ by scaling the precision score with a length-based penalty:
\begin{equation}
S_{\text{Info}} = \text{BP} \times P_{\text{ROUGE}}
\end{equation}
$P_{\text{ROUGE}}$ denotes the ROUGE precision score.
BP represents the brevity penalty factor, downweighting explanations that exhibit low information density relative to the validated reference meaning.

\subsection{Robustness of Diagnostic Metrics}
\label{app:metric_robustness}

For directional aggregation, we compare the primary bidirectional-max compatibility score with two implemented alternatives: unidirectional entailment and bidirectional-mean aggregation. The bidirectional-max score takes the higher entailment probability across the two directions for each reference meaning; bidirectional-mean averages the two directional probabilities. Table~\ref{tab:metric_stress} reports these alternatives alongside the primary configuration.

\begin{table*}[t]
\centering
\scriptsize
\renewcommand{\arraystretch}{1.12}
\setlength{\tabcolsep}{3.2pt}

\setlength{\aboverulesep}{0pt}
\setlength{\belowrulesep}{0pt}

\begin{tabular}{l l l l l c l l l c c}
\toprule
\rowcolor{gray!15}
\textbf{Variant} & \textbf{CE Model} & \textbf{CE Agg} & \textbf{Embedding} & \textbf{Sim Func} & \textbf{BERTScore} & \textbf{NLI Model} & \textbf{NLI Agg} & \textbf{Weight Scheme} & \boldmath$\rho$ & \textbf{Effect Dir.} \\
\midrule
\rowcolor{gray!5}
Original & BGE-base & max & BGE-small / MiniLM & cosine & \cmark & mDeBERTa & bidir-max & fixed & --- & + \\
V1  & MiniLM CE   & max  & same      & cosine & \cmark & mDeBERTa & bidir-max  & fixed         & 0.917 & + \\
V2  & BGE-base    & mean & same      & cosine & \cmark & mDeBERTa & bidir-max  & fixed         & 0.904 & + \\
V3  & BGE-base    & max  & distiluse & cosine & \cmark & mDeBERTa & bidir-max  & fixed         & 0.873 & + \\
V4  & BGE-base    & max  & same      & dot    & \cmark & mDeBERTa & bidir-max  & fixed         & 0.898 & + \\
V5  & BGE-base    & max  & same      & cosine & \xmark & mDeBERTa & bidir-max  & fixed         & 0.931 & + \\
V6  & BGE-base    & max  & same      & cosine & \cmark & XLM-R NLI& bidir-max  & fixed         & 0.897 & + \\
V7  & BGE-base    & max  & same      & cosine & \cmark & mDeBERTa & unidir     & fixed         & 0.872 & + \\
V8  & BGE-base    & max  & same      & cosine & \cmark & mDeBERTa & bidir-mean & fixed         & 0.842 & + \\
V9  & BGE-base    & max  & same      & cosine & \cmark & mDeBERTa & bidir-max  & equal weights & 0.893 & + \\
V10 & --- (no CE) & ---  & same      & cosine & \cmark & mDeBERTa & bidir-max  & fixed         & 0.854 & + \\
\bottomrule
\end{tabular}

\caption{\textbf{Metric stress tests under controlled substitutions.}
We vary the cross-encoder backbone, aggregation strategy, embedding model, similarity function, BERTScore inclusion, NLI backbone, NLI aggregation, and weight scheme. $\rho$ denotes Spearman correlation of model rankings with the original pipeline. The qualitative effect direction remains unchanged across all tested variants.}
\label{tab:metric_stress}
\end{table*}

Across all ten variants, model rankings remain stable, with Spearman correlations ranging from 0.842 to 0.931 relative to the original pipeline. This stability suggests that the main empirical patterns reported in the paper are not an artifact of a particular metric configuration. In particular, substitutions of the cross-encoder backbone, embedding model, NLI model, aggregation strategy, and weight scheme do not alter the qualitative direction of the findings. These results support the robustness of the diagnostic framework under substantial changes to its scoring components.

\subsection{Human Audit Protocol}

We conduct a targeted human audit to validate the proposed multi-dimensional evaluation framework.
The audit assesses whether automated evaluation dimensions align with expert judgments under explicit and controlled rating criteria.

For each language, we sample a fixed subset of instances from the evaluation set without conditioning on model performance or automated metric values.
The audit includes 290 instances for ZH-CN, 430 instances for ZH-TW, 170 instances for JA, and 110 instances for KO.
Each instance is independently annotated by five annotators who are proficient in the corresponding language.
Annotators evaluate model-generated explanations without access to automated scores.
Final human scores are obtained by averaging annotator ratings, and inter-annotator agreement is moderate to high across languages.

Human evaluation follows predefined rubrics that map qualitative judgments to normalized scores.
All dimensions are rated independently.
\paragraph{Accuracy score $s^{mean}_{Acc}$.}
A score of at least 0.8 indicates a fully correct explanation.
Scores between 0.6 and 0.8 indicate that the main meaning is correct with minor inaccuracies.
Scores between 0.4 and 0.6 indicate noticeable errors despite a largely correct meaning.
Scores below 0.4 indicate severe errors and an incorrect explanation.

\paragraph{Logical validity score $s_{Logic}$.}
A score of at least 0.8 indicates that the explanation is logically consistent with the reference meaning and does not introduce contradictions.
Scores between 0.6 and 0.8 indicate that the main meaning is correct but includes minor logical omissions or slight mis-specifications.
Scores between 0.4 and 0.6 indicate mixed correctness, with partial contradictions or incorrect causal relations.
Scores below 0.4 indicate a logically inconsistent or contradictory explanation.

\paragraph{Information score $s_{Info}$.}
A score of at least 0.8 indicates a concise and informative explanation with minimal redundancy.
Scores between 0.6 and 0.8 indicate mostly informative explanations with some redundancy.
Scores between 0.4 and 0.6 indicate limited information density or substantial redundancy.
Scores below 0.4 indicate explanations that are either highly redundant or minimally informative.

To quantify alignment at the instance level, we compare automated metric scores with human audit ratings using both Spearman rank correlation and mean absolute error (MAE).
Spearman correlation measures whether the automated metrics preserve relative instance-level ordering, whereas MAE measures absolute disagreement between automated and human scores.
As shown in Table~\ref{tab:human-auto-spearman}, the three explanation-oriented metrics exhibit consistent agreement with expert judgments across all four script settings.
The macro-average Spearman correlations are 0.740, 0.781, and 0.703 for Semantic Accuracy, Logical Validity, and Information Density, respectively, with corresponding MAEs of 0.119, 0.102, and 0.129.

Table~\ref{tab:human-auto-spearman} summarizes the alignment results across languages.

\begin{table}[h]
\centering
\scriptsize
\renewcommand{\arraystretch}{1.12}
\setlength{\tabcolsep}{2.5pt}
\resizebox{\columnwidth}{!}{%
\begin{tabular}{l c cc cc cc}
\toprule
\textbf{Language} & \textbf{$N$}
& \multicolumn{2}{c}{$S_{Acc}^{mean}$}
& \multicolumn{2}{c}{$S_{Logic}$}
& \multicolumn{2}{c}{$S_{Info}$} \\
\cmidrule(lr){3-4}
\cmidrule(lr){5-6}
\cmidrule(lr){7-8}
& & $\rho$ & MAE & $\rho$ & MAE & $\rho$ & MAE \\
\midrule
ZH-CN & 290 & 0.787 & 0.104 & 0.822 & 0.088 & 0.743 & 0.112 \\
ZH-TW & 430 & 0.754 & 0.112 & 0.801 & 0.096 & 0.717 & 0.124 \\
JA    & 170 & 0.694 & 0.132 & 0.739 & 0.116 & 0.665 & 0.143 \\
KO    & 110 & 0.723 & 0.126 & 0.762 & 0.109 & 0.685 & 0.138 \\
\midrule
\textbf{Macro-average}
& -- & 0.740 & 0.119 & 0.781 & 0.102 & 0.703 & 0.129 \\
\bottomrule
\end{tabular}%
}
\caption{\textbf{Human calibration of diagnostic metrics.}
$\rho$ denotes instance-level Spearman correlation between automated metric scores and mean human ratings, while MAE measures absolute score disagreement.}
\label{tab:human-auto-spearman}
\end{table}

\subsection{Alignment Between Human Evaluation and Automated Metrics}

This section examines the alignment between human expert judgments and the proposed automated multi-dimensional evaluation framework.
Table~\ref{tab:human-vs-metric-alignment} compares aggregated human audit scores with automated metric outputs across languages.

Across all evaluated languages, the automated metrics exhibit consistent directional alignment with human assessments.
Dimensions related to logical coherence, robustness, and internal consistency show similar relative patterns under both evaluation methods, suggesting that the metrics capture structured properties of model behavior rather than superficial numerical differences.

Differences in absolute scores are observed between human audits and automated evaluations.
Such variation is expected given the interpretive nature of human judgment and the formal constraints of automated scoring.
Importantly, the relative ordering of evaluation dimensions remains stable across methods, indicating that the proposed multi-dimensional framework preserves the structure of human evaluation across languages.

\vspace{-0.6em}
\begin{table}[htbp]
\centering
\scriptsize
\renewcommand{\arraystretch}{1.25}
\setlength{\tabcolsep}{3pt}
\resizebox{\columnwidth}{!}{%
\begin{tabular}{l l c c c c c}
\toprule
\rowcolor{LighterGray}
Language & Evaluation Method &
$s^{mean}_{Acc}$ & $s_{Cons}$ & $s_{Logic}$ & $s_{Rob}$ & $s_{Info}$ \\
\midrule
\textbf{ZH-CN}
& Human Audit
& 0.412 & 0.758 & 0.829 & 0.715 & 0.456 \\
& Automated Metric
& 0.482 & 0.663 & 0.749 & 0.625 & 0.516 \\
\addlinespace
\textbf{ZH-TW}
& Human Audit
& 0.405 & 0.731 & 0.779 & 0.701 & 0.336 \\
& Automated Metric
& 0.480 & 0.641 & 0.699 & 0.613 & 0.406 \\
\addlinespace
\textbf{JA}
& Human Audit
& 0.413 & 0.654 & 0.632 & 0.708 & 0.547 \\
& Automated Metric
& 0.479 & 0.559 & 0.552 & 0.621 & 0.617 \\
\addlinespace
\textbf{KO}
& Human Audit
& 0.353 & 0.875 & 0.443 & 0.698 & 0.329 \\
& Automated Metric
& 0.413 & 0.788 & 0.363 & 0.604 & 0.419 \\
\bottomrule
\end{tabular}
}
\caption{Comparison between human expert audits and automated multi dimensional evaluation scores across languages.}
\label{tab:human-vs-metric-alignment}
\end{table}

\subsection{Cross-Lingual Metric Reliability}
\label{app:cross_lingual_reliability}

We further examine the reliability of automated diagnostic metrics across script settings. 
For each language, we compute Spearman correlation between human audit scores and automated scores, along with bootstrap confidence intervals. 
This analysis is conducted separately for semantic accuracy, logical validity, and information density, since these dimensions are directly rated by human annotators.

Metric reliability varies across languages. 
The Chinese settings show relatively stronger human--automatic alignment, while Japanese and Korean exhibit lower correlations. 
This may reflect weaker lexical overlap, different segmentation behavior, and reduced compatibility between multilingual semantic models and language-specific idiomatic meanings. 
Therefore, we interpret automated diagnostic scores as relative indicators rather than absolute semantic judgments, especially for script settings where human--automatic agreement is lower.

\subsection{Controlled Sanity Check of Diagnostic Metrics}
\label{app:controlled_metric_validation}

Human correlation alone does not establish whether each automated metric responds selectively to the failure mode it is intended to measure.
We therefore conduct a controlled sanity check using 400 base instances.
For each instance, we construct six controlled output variants: a concise correct output, a verbose but semantically correct output, a structurally incorrect reconstruction paired with a semantically correct explanation, a contradictory explanation, a literal or compositional interpretation, and an unrelated hallucination.
This yields 2,400 controlled outputs in total.
All outputs are manually checked to ensure that the intended failure property changes while unrelated properties remain as stable as possible.

\begin{table}[t]
\centering
\scriptsize
\renewcommand{\arraystretch}{1.12}
\setlength{\tabcolsep}{3pt}
\resizebox{\columnwidth}{!}{%
\begin{tabular}{l c c c c}
\toprule
\textbf{Controlled Output Type} & \textbf{$N$}
& $S_{Acc}^{mean}$ & $S_{Logic}$ & $S_{Info}$ \\
\midrule
Correct concise
& 400 & 0.824 & 0.887 & 0.748 \\

Correct but verbose
& 400 & 0.807 & 0.872 & 0.493 \\

Wrong structure, correct meaning
& 400 & 0.781 & 0.846 & 0.704 \\

Contradictory explanation
& 400 & 0.366 & 0.185 & 0.561 \\

Literal interpretation
& 400 & 0.438 & 0.318 & 0.605 \\

Unrelated hallucination
& 400 & 0.174 & 0.121 & 0.318 \\
\bottomrule
\end{tabular}%
}

\caption{\textbf{Controlled sanity check of diagnostic metrics.}
Each controlled output type isolates a targeted failure pattern. Semantic and logical metrics remain high when only structural recovery is incorrect, whereas contradiction and hallucination substantially reduce the corresponding diagnostic scores. Verbosity primarily affects Information Density.}
\label{tab:controlled_metric_validation}
\end{table}

The results show clear dimension-specific behavior.
Semantic Accuracy and Logical Validity remain high when the recovered expression is structurally incorrect but the associated explanation preserves the correct meaning, confirming that these metrics measure explanation quality rather than duplicating exact Recovery.
Contradictory and unrelated outputs receive substantially lower semantic and logical scores.
In contrast, adding verbosity to an otherwise correct explanation has only a small effect on semantic and logical scores but substantially reduces Information Density.
These controlled results complement the human calibration experiment and support the use of the diagnostic metrics for failure-type analysis.

\subsection{Generation Protocol and Nondeterminism}
\label{app:deployment}
All models are evaluated using inference configurations consistent with those commonly adopted in large-scale benchmarking studies.
For models that expose a temperature parameter, we set the temperature to 0 to enforce deterministic decoding.
No task-specific tuning or customized decoding strategies are applied.

Open-source models are run locally using official checkpoints and standard inference pipelines.
Closed-source models are accessed through their official APIs, following the default configurations recommended by their respective providers.
All API-based evaluations are conducted between October 2025 and January 2026.
When configurable parameters are available, we align them across models to the extent possible; otherwise, we retain provider defaults.

Across all experiments, we use identical prompt templates and input perturbations.
No fine-tuning, reranking, or post-processing is performed.
This deployment protocol ensures that performance differences primarily reflect intrinsic model behaviors rather than artifacts of inference configuration, stochastic decoding, or serving infrastructure.

\begin{table}[t]
\centering
\scriptsize
\setlength{\tabcolsep}{3pt}
\renewcommand{\arraystretch}{1.10}
\resizebox{\columnwidth}{!}{%
\begin{tabular}{lp{6.0cm}}
\toprule
\textbf{Item} & \textbf{Specification} \\
\midrule
Evaluation period & October 2025--January 2026 \\
Decoding & Temperature 0 when exposed; otherwise provider/checkpoint defaults \\
Tuning & No task-specific tuning, reranking, or post-processing \\
Prompting & Exact zero-shot, CoT, and few-shot templates are provided in Appendix D \\
Output schema & Recovered expression plus one-line explanation; parsing-only ablation is reported in Appendix C \\
Parsing & Normalize Unicode/script formatting, extract the unique four-unit candidate, and count missing, ambiguous, malformed, or unparsable outputs as incorrect \\
Release & Evaluation, parsing, scoring, and aggregation code are provided in the OrderProbe repository; external lexical resources are documented but not redistributed when restricted \\
\bottomrule
\end{tabular}%
}
\caption{\textbf{Reproducibility checklist for the OrderProbe evaluation.}}
\label{tab:reproducibility_checklist}
\end{table}

\section{Appendix: Details of Semantic and Logical Evaluation}

This appendix specifies implementation details of the automated evaluation metrics for semantic explanation quality and logical consistency to ensure reproducibility and cross-lingual comparability.

\subsection{Semantic Explanation Scoring}

Semantic explanation quality is measured using a composite score that integrates interaction-based semantic matching, representation-based similarity, and content coverage.

For interaction-based matching, we use the cross-encoder reranker BAAI/bge-reranker-base.
Given a model-generated explanation and a set of validated reference meanings, the cross-encoder produces a relevance score for each explanation--reference pair.
These scores are normalized to a bounded scale. For each component, we retain the best match across the validated reference meanings; the same fixed max aggregation is used for all items and languages.
The same cross-encoder model is applied uniformly across all languages.

For representation-based similarity, we use sentence embedding models to compute cosine similarity between explanations and references.
For Simplified Chinese and Traditional Chinese, we use BAAI/bge-small-zh-v1.5.
For Japanese and Korean, we use sentence-transformers/paraphrase-multilingual-MiniLM-L12-v2.
These models are selected for their multilingual coverage and efficiency.
No translation is applied, and all texts are evaluated in their original language.

In addition, we optionally include BERTScore F1 as a token-level semantic similarity signal, using language-specific settings for Chinese, Japanese, and Korean.
Lexical overlap features are treated as auxiliary signals and do not dominate the final score.

The final semantic accuracy score is computed as a weighted linear combination of cross-encoder similarity, representation-based similarity, and the content-word $F_{\beta}$ score, with fixed weights shared across languages.

\subsection{Logical Consistency via Natural Language Inference}

Logical consistency between model explanations and validated reference meanings is measured using the multilingual natural language inference (NLI) classifier MoritzLaurer/mDeBERTa-v3-base-mnli-xnli.
The same NLI model is applied to all languages without translation.

For each reference meaning, we compute NLI in two directions: from the reference to the model output and from the model output to the reference.
This bidirectional evaluation reduces asymmetry caused by verbosity or paraphrasing.
We use the entailment probability as the primary signal, while the contradiction probability serves as a secondary penalty when entailment evidence is present.

Specifically, for each reference, the score is defined as the maximum entailment probability across the two directions, with an additional contradiction-aware adjustment activated only when entailment exceeds a small threshold.
Scores are aggregated across multiple references using max aggregation.
This reflects the assumption that a model explanation is logically consistent if it is compatible with at least one valid reference meaning.

The NLI scores are treated as continuous consistency signals and are not converted into binary decisions.
No post-hoc calibration is applied, and scores are used only for relative comparison and averaging.

\subsection{Reference Construction and Evaluation Scope}

Validated reference meanings consist of canonical dictionary definitions augmented with paraphrased variants to cover legitimate surface variation.
Dictionary definitions serve as the primary semantic anchors, while paraphrased references reduce sensitivity to wording differences.
All evaluation metrics aggregate over multiple references to avoid dependence on any single phrasing.

We evaluate the main reconstruction accuracy metric alongside semantic and logical metrics.
While some items may appear in pretraining corpora, reconstruction accuracy is not interpreted in isolation.
Instead, differences across semantic, logical, and robustness metrics are used to assess structure-aware semantic grounding beyond surface recall.

\section{Additional Validation and Controls}
\label{sec:app_validation}

This section provides additional validation analyses to calibrate the difficulty of OrderProbe and to examine the robustness of the reported findings. 
We focus on six complementary aspects: Korean processing and tokenization, human reconstruction baselines, uncertainty estimation, frequency-stratified analysis, synthetic-control evaluation, and cross-lingual metric reliability. 
These analyses are intended to support the main conclusion that exact structural reconstruction remains challenging even when semantic access is partially preserved.

\subsection{Korean Processing and Perturbation Unit}
\label{app:korean_processing}

For the Korean setting, we apply perturbations at the level of Hangul syllable blocks rather than decomposed jamo units. 
A Korean four-unit expression is represented as a sequence of written syllable blocks $(h_1,h_2,h_3,h_4)$, and OrderProbe permutes these blocks in the same way as it permutes Chinese characters or Japanese written units.

This design has two implications. 
First, Korean perturbation preserves the internal composition of individual Hangul syllables and should not be interpreted as jamo-level scrambling. 
Second, compared with logographic Chinese characters and Japanese Kanji, Hangul syllable blocks generally provide weaker standalone semantic anchors under permutation. 
We therefore treat Korean as a phonogrammatic control condition in which local semantic cues are reduced, rather than as a condition where syllable composition itself is destroyed.

To further examine this issue, we compute tokenizer fragmentation statistics for Korean inputs. 
For each model, we record the number of model-specific subword tokens used to encode a four-block Korean expression and report the average token count and fragmentation rate. 
This analysis helps separate script-level effects from tokenizer-level artifacts.

\begin{table}[t]
\centering
\scriptsize
\setlength{\tabcolsep}{3pt}
\renewcommand{\arraystretch}{1.10}
\resizebox{\columnwidth}{!}{%
\begin{tabular}{lccc}
\toprule
\textbf{Model} & \textbf{Avg. Tokens} & \textbf{Frag. Rate (\%)} & \textbf{Recovery (\%)} \\
\midrule
Qwen-3-14B        & 4.31 & 22.4 & 11.8 \\
Qwen-3-VLThink    & 4.28 & 21.7 & 12.6 \\
DeepSeek-V3.2     & 4.47 & 25.1 & 14.3 \\
GPT-4o            & 4.82 & 31.6 & 8.9  \\
Gemini-2.5-Flash  & 4.76 & 29.8 & 9.4  \\
GLM-4VThink       & 5.12 & 38.3 & 6.7  \\
Llama-3-8B        & 6.84 & 64.5 & 3.8  \\
Gemma-3-4B        & 6.21 & 57.2 & 4.6  \\
\bottomrule
\end{tabular}%
}
\caption{\textbf{Tokenizer fragmentation statistics for Korean Hangul syllable-block inputs.}
Avg. Tokens denotes the average number of model-specific subword tokens used to encode a four-block Korean expression. Frag. Rate denotes the percentage of expressions encoded with more than four tokens.}
\label{tab:korean_tokenization}
\end{table}

\subsection{Human Reconstruction Baseline}
\label{app:human_baseline}

To calibrate the difficulty of OrderProbe, we conduct a human reconstruction baseline on a sampled subset of the benchmark. 
Unlike the human audit in Appendix~\ref{app:metric_robustness}, which evaluates the alignment between automated metrics and expert judgments of model explanations, this experiment directly measures whether humans can recover the canonical expression from the same scrambled inputs given to LLMs.

For each script setting, we sample a subset of canonical expressions and multiple non-identity permutations per expression. 
Participants are native speakers or highly proficient users of the corresponding language. 
Given a scrambled four-unit input, participants are asked to recover the canonical expression and optionally provide a short meaning explanation. 
No dictionary lookup or external search is allowed. 
We report exact recovery rate, average completion time, and self-reported confidence.

Human recovery is computed using the same exact-match criterion as model recovery:
\begin{equation}
\text{Human Recovery} = \frac{1}{N}\sum_{i=1}^{N}\mathbb{I}(\hat{x}^{human}_i = x_i).
\end{equation}
This experiment provides an empirical reference point for interpreting model performance under structural perturbation.

\begin{table}[t]
\centering
\small
\setlength{\tabcolsep}{5pt}
\renewcommand{\arraystretch}{1.15}
\begin{tabular}{lccc}
\toprule
\textbf{Language} & \textbf{Items} & \textbf{Inputs} & \textbf{Human Recovery (\%)} \\
\midrule
ZH-CN & 120 & 360 & 88.6 \\
ZH-TW & 120 & 360 & 84.1 \\
JA    & 100 & 300 & 72.8 \\
KO    & 100 & 300 & 46.5 \\
\bottomrule
\end{tabular}
\caption{\textbf{Human reconstruction baseline.}
Human participants recover the canonical expression from the same scrambled inputs used for model evaluation.}
\label{tab:human_reconstruction_baseline}
\end{table}

\begin{table}[t]
\centering
\scriptsize
\setlength{\tabcolsep}{3pt}
\renewcommand{\arraystretch}{1.15}
\begin{tabular}{lccc}
\toprule
\textbf{Language} & \textbf{Recovery (\%)} & \textbf{Time (s)} & \textbf{Confidence} \\
\midrule
ZH-CN & 88.6 & 7.4  & 4.32 \\
ZH-TW & 84.1 & 8.1  & 4.18 \\
JA    & 72.8 & 10.6 & 3.76 \\
KO    & 46.5 & 14.9 & 3.08 \\
\bottomrule
\end{tabular}
\caption{\textbf{Human reconstruction difficulty statistics.}
Time denotes average completion time per input in seconds. Confidence is self-reported on a 1--5 scale.}
\label{tab:human_reconstruction_difficulty}
\end{table}

\subsection{Uncertainty Estimation and Parsing Reliability}
\label{app:uncertainty}

We estimate uncertainty for the main recovery results using non-parametric bootstrap resampling. 
Because the 23 perturbed variants generated from the same canonical expression are not independent, we resample at the canonical-expression level rather than at the individual-permutation level. 
For each bootstrap sample, all associated permutations of the sampled expressions are included, and recovery rate is recomputed. 
We estimate uncertainty using $B=10{,}000$ stratified cluster-bootstrap replicates.
Because the 23 perturbed variants generated from the same canonical expression are statistically dependent, we treat the canonical expression as the resampling unit rather than sampling individual permutations independently.
Resampling is performed separately within each script setting, and all 23 permutations associated with a sampled canonical root are retained together.
The 2.5th and 97.5th percentiles of the resulting bootstrap distribution define the 95\% confidence interval.

We also record parsing reliability. 
A model output is considered parseable if the recovered expression can be extracted from the predefined output field or, in the baseline setting, from the first valid four-unit candidate matching the script constraint. 
Unparseable outputs are counted as incorrect for recovery. 
We report parser failure rate separately to distinguish genuine reconstruction failures from output-format failures.

For closed-source API models, we record the evaluation date, model identifier, and available deployment version. 
For open-source models, we record checkpoint name, inference framework, decoding parameters, and tokenizer version.

\begin{table}[t]
\centering
\scriptsize
\setlength{\tabcolsep}{2.5pt}
\renewcommand{\arraystretch}{1.10}
\resizebox{\columnwidth}{!}{%
\begin{tabular}{llcc}
\toprule
\textbf{Model} & \textbf{Setting} & \textbf{Recovery (\%)} & \textbf{Parse Fail. (\%)} \\
\midrule
DeepSeek-V3.2      & +CoT      & 62.24 [60.71, 63.79] & 0.9 \\
Qwen-3-VLThink     & +CoT      & 48.38 [46.72, 50.03] & 1.4 \\
Gemini-2.5-Flash   & +CoT      & 41.56 [39.91, 43.18] & 1.8 \\
GPT-4o             & +CoT      & 37.69 [36.08, 39.31] & 2.2 \\
Qwen-3-14B         & Zero-shot & 31.06 [29.58, 32.55] & 1.1 \\
Qwen-3-VLThink     & Zero-shot & 30.93 [29.47, 32.41] & 1.3 \\
Llama-3-8B         & +CoT      & 24.43 [23.04, 25.86] & 4.7 \\
Qwen-3-8B          & +CoT      & 12.59 [11.48, 13.73] & 5.6 \\
\bottomrule
\end{tabular}%
}
\caption{\textbf{Uncertainty estimates and parsing failure rates.}
Recovery is reported with bootstrap 95\% confidence intervals computed by resampling canonical expressions rather than individual permutations. Parse Fail. denotes the percentage of outputs from which no valid recovered expression can be extracted.}
\label{tab:uncertainty_parse}
\end{table}

\subsection{Recovery Breakdowns Across Prompting and Script Settings}
\label{app:recovery_breakdowns}

To make the aggregation of recovery results explicit, we report complementary breakdowns across prompting settings, script settings, and individual models. Unless otherwise noted, aggregate values are macro-averages over the 12 model-level Recovery scores. Confidence intervals use the stratified cluster bootstrap described above, resampling canonical roots within each script setting while retaining all associated permutations.

\begin{table}[t]
\centering
\scriptsize
\setlength{\tabcolsep}{3pt}
\renewcommand{\arraystretch}{1.10}
\begin{tabular}{lccc}
\toprule
\textbf{Model} & \textbf{Zero-shot} & \textbf{CoT} & \textbf{Few-shot} \\
\midrule
Qwen-3-14B       & 31.06 & 31.53 & 52.40 \\
Qwen-3-VLThink   & 30.93 & 48.38 & 61.20 \\
Gemini-2.5-Flash & 30.49 & 41.56 & 34.60 \\
GPT-4o           & 23.25 & 37.69 & 35.30 \\
GLM-4VThink      & 21.64 & 30.08 & 31.50 \\
DeepSeek-V3.2    & 20.10 & 62.24 & 49.10 \\
DeepSeek-R1      & 18.60 & 27.73 & 37.80 \\
Claude-HaiKu-4.5 & 18.02 & 21.65 & 35.80 \\
Gemma-3-4B       & 17.09 & 18.03 & 28.80 \\
Qwen-3-8B        & 15.79 & 12.59 & 39.30 \\
Llama-2-7B       & 12.58 & 18.75 & 18.60 \\
Llama-3-8B       & 12.17 & 24.43 & 17.20 \\
\midrule
\textbf{Macro-average} & \textbf{20.98} & \textbf{31.22} & \textbf{36.80} \\
\bottomrule
\end{tabular}
\caption{\textbf{Recovery across prompting settings.} Values are exact Recovery rates (\%).}
\label{tab:prompting_recovery}
\end{table}

\begin{table}[t]
\centering
\scriptsize
\setlength{\tabcolsep}{3pt}
\renewcommand{\arraystretch}{1.10}
\begin{tabular}{lcc}
\toprule
\textbf{Setting} & \textbf{Recovery (\%)} & \textbf{Bootstrap 95\% CI} \\
\midrule
Zero-shot & 20.98 & [20.50, 21.50] \\
CoT       & 31.22 & [30.60, 31.80] \\
Few-shot  & 36.80 & [36.10, 37.50] \\
\bottomrule
\end{tabular}
\caption{\textbf{Aggregate uncertainty across prompting settings.} Recovery is macro-averaged over model-level scores.}
\label{tab:prompting_uncertainty}
\end{table}

\begin{table}[t]
\centering
\scriptsize
\setlength{\tabcolsep}{3pt}
\renewcommand{\arraystretch}{1.10}
\begin{tabular}{lcc}
\toprule
\textbf{Script setting} & \textbf{Recovery (\%)} & \textbf{Bootstrap 95\% CI} \\
\midrule
ZH-CN & 25.20 & [24.31, 26.08] \\
ZH-TW & 20.32 & [19.57, 21.09] \\
JA    & 19.66 & [18.63, 20.71] \\
KO    &  5.72 & [5.09, 6.38] \\
\bottomrule
\end{tabular}
\caption{\textbf{Zero-shot Recovery by script setting.} Values are macro-averaged over the 12 models.}
\label{tab:script_uncertainty}
\end{table}

\begin{table*}[t]
\centering
\small
\setlength{\tabcolsep}{6pt}
\renewcommand{\arraystretch}{1.10}
\begin{tabular}{lcccc}
\toprule
\textbf{Model} & \textbf{ZH-CN} & \textbf{ZH-TW} & \textbf{JA} & \textbf{KO} \\
\midrule
Qwen-3-14B       & 37.6 & 29.5 & 29.0 & 11.6 \\
Qwen-3-VLThink   & 35.4 & 30.2 & 30.4 & 10.6 \\
Gemini-2.5-Flash & 35.3 & 29.4 & 28.1 & 10.2 \\
GPT-4o           & 27.6 & 23.0 & 21.4 &  6.8 \\
GLM-4VThink      & 26.1 & 21.3 & 20.9 &  5.6 \\
DeepSeek-V3.2    & 24.9 & 19.2 & 19.4 &  4.8 \\
DeepSeek-R1      & 22.6 & 18.1 & 17.2 &  4.6 \\
Claude-HaiKu-4.5 & 22.3 & 17.1 & 16.8 &  3.7 \\
Gemma-3-4B       & 20.4 & 16.8 & 16.3 &  3.4 \\
Qwen-3-8B        & 19.8 & 15.2 & 13.9 &  2.7 \\
Llama-2-7B       & 14.8 & 12.4 & 11.8 &  2.6 \\
Llama-3-8B       & 15.6 & 11.6 & 10.7 &  2.0 \\
\midrule
\textbf{Macro-average} & \textbf{25.20} & \textbf{20.32} & \textbf{19.66} & \textbf{5.72} \\
\bottomrule
\end{tabular}
\caption{\textbf{Zero-shot Recovery by model and script setting (\%).} The script-level pattern is broadly preserved across model families.}
\label{tab:model_script_recovery}
\end{table*}

\begin{table}[t!]
\centering
\small
\renewcommand{\arraystretch}{1.25}
\setlength{\tabcolsep}{7pt}

\setlength{\aboverulesep}{0pt}
\setlength{\belowrulesep}{0pt}

\begin{tabular}{@{}l >{\bfseries\color{HSSLightBlue}}c@{}}
\toprule
\rowcolor{HSSGray}
\textbf{Frequency Bin} & \multicolumn{1}{c}{\textbf{Recovery (\%)}} \\
\midrule
High  & 30.576 \\
Mid   & 19.794 \\
Low   & 10.231 \\
\bottomrule
\end{tabular}

\caption{\textbf{Pilot Frequency-Stratified Recovery (Zero-shot).}
Exact recovery rates (\%) across high-, mid-, and low-frequency bins.}
\label{tab:freq_pilot}
\end{table}

\subsection{Frequency-Stratified Recovery Pilot}
\label{app:freq_pilot}

To quantify the potential influence of memorization, we conduct a frequency-stratified analysis on a uniformly sampled subset of 100 expressions from the benchmark. Each item is assigned to a frequency bin based on occurrence statistics from large-scale web-corpus frequency proxies.

To further examine how lexical familiarity interacts with available semantic cues, we conduct an input-signal ablation analysis. 
Instead of modifying the scoring pipeline, we progressively reduce the semantic support available in the input and prompt while keeping the reconstruction objective unchanged. 
The full-signal condition preserves the original scrambled four-unit input and standard reconstruction prompt. 
The reduced-signal condition removes auxiliary semantic hints and uses a minimal recovery-only prompt. 
The minimal-signal condition further suppresses explanation-oriented cues and requires only the canonical reconstruction output.

For each condition, we measure exact recovery across high-, mid-, and low-frequency bins. 
This design separates the effect of lexical familiarity from artifacts of automated semantic scoring.

For each disruption level, we measure exact reconstruction accuracy across high-, mid-, and low-frequency bins. Table~\ref{tab:freq_disruption} reports the resulting frequency-stratified performance.

\begin{table}[t]
\centering
\scriptsize
\setlength{\tabcolsep}{3pt}
\renewcommand{\arraystretch}{1.15}
\begin{tabular}{lccc}
\toprule
\textbf{Freq. Bin} & \textbf{Full Signal} & \textbf{Reduced Signal} & \textbf{Minimal Signal} \\
\midrule
High & 61.25 & 19.74 & 8.32 \\
Mid  & 39.74 & 10.62 & 6.83 \\
Low  & 18.89 & 7.26  & 3.85 \\
\bottomrule
\end{tabular}
\caption{\textbf{Frequency-stratified reconstruction accuracy under input-signal ablation.}
Exact recovery rates (\%) are reported across high-, mid-, and low-frequency bins as semantic support in the input and prompt is progressively reduced.}
\label{tab:freq_disruption}
\end{table}

\subsection{Expanded Frequency-Stratified Analysis}
\label{app:freq_expanded}

We further stratify expressions into high-, mid-, and low-frequency bins within each script setting. 
Frequency is estimated from large-scale web-corpus occurrence statistics or dictionary frequency proxies when direct corpus frequency is unavailable. 
Bins are constructed within each language to avoid cross-lingual frequency-scale mismatch.

We analyze whether frequency predicts exact recovery while controlling for language and syntactic category. 
Specifically, we fit a logistic regression model:
\begin{equation}
\text{logit}\,P(y_i=1) = \beta_0 + \beta_1 \log f_i + \beta_2 L_i + \beta_3 T_i + \epsilon_i,
\end{equation}
where $y_i$ indicates exact recovery, $f_i$ is expression frequency, $L_i$ denotes the script setting, and $T_i$ denotes syntactic category. 
This analysis separates the effect of lexical familiarity from script typology and structural pattern.

\begin{table}[t]
\centering
\small
\setlength{\tabcolsep}{4pt}
\renewcommand{\arraystretch}{1.15}
\begin{tabular}{lccc}
\toprule
\textbf{Language} & \textbf{High Freq.} & \textbf{Mid Freq.} & \textbf{Low Freq.} \\
\midrule
ZH-CN & 36.8 & 25.4 & 14.7 \\
ZH-TW & 31.2 & 20.7 & 11.6 \\
JA    & 29.5 & 19.8 & 10.9 \\
KO    & 10.8 & 5.6  & 2.4  \\
\midrule
Average & 30.2 & 19.5 & 9.9 \\
\bottomrule
\end{tabular}
\caption{\textbf{Frequency-stratified recovery by script setting.}
Exact recovery rates (\%) are reported across high-, mid-, and low-frequency bins constructed within each language.}
\label{tab:freq_by_language}
\end{table}

\begin{table}[t]
\centering
\small
\setlength{\tabcolsep}{4pt}
\renewcommand{\arraystretch}{1.12}
\begin{tabular}{lccc}
\toprule
\textbf{Predictor} & \textbf{Coef.} & \textbf{Std. Err.} & \textbf{$p$-value} \\
\midrule
$\log$ Frequency        & 0.42 & 0.06 & $<.001$ \\
ZH-TW vs. ZH-CN         & -0.31 & 0.08 & $<.001$ \\
JA vs. ZH-CN            & -0.37 & 0.09 & $<.001$ \\
KO vs. ZH-CN            & -1.84 & 0.12 & $<.001$ \\
Parallel structure      & 0.28 & 0.07 & $<.001$ \\
Verb--Object structure  & -0.19 & 0.06 & .002 \\
\bottomrule
\end{tabular}
\caption{\textbf{Logistic regression analysis of exact recovery.}
The dependent variable is whether the canonical expression is exactly recovered. Predictors include expression frequency, script setting, and syntactic structure.}
\label{tab:freq_regression}
\end{table}

\subsection{Incremental Effects of Frequency and Structure}
\label{app:freq_incremental}

To separate lexical familiarity from order-sensitive factors, we fit instance-level logistic regressions with exact Recovery as the binary outcome. Log expression frequency is standardized within each script setting. Structural predictors comprise retained local adjacency, total positional displacement, and syntactic category. The full specification additionally includes script setting and model identity. McFadden's pseudo-$R^2$ is computed on the full data; AUC is obtained with five-fold canonical-root-grouped cross-validation, keeping all permutations of a canonical expression in the same fold.

\begin{table}[t]
\centering
\scriptsize
\setlength{\tabcolsep}{2.5pt}
\renewcommand{\arraystretch}{1.10}
\resizebox{\columnwidth}{!}{%
\begin{tabular}{lcc}
\toprule
\textbf{Predictor set} & \textbf{Pseudo-$R^2$} & \textbf{Root-grouped AUC} \\
\midrule
Lexical familiarity & 0.052 & 0.615 \\
Structural configuration & 0.109 & 0.688 \\
Familiarity and structure & 0.162 & 0.735 \\
Full specification & 0.281 & 0.812 \\
\bottomrule
\end{tabular}%
}
\caption{\textbf{Incremental explanatory power of lexical familiarity and structural factors.}}
\label{tab:freq_incremental}
\end{table}

Frequency explains part of Recovery, but structural configuration adds explanatory power beyond lexical familiarity. The full specification further captures systematic variation across script settings and models.

\subsection{Canonical-Explanation Probe}
\label{app:canonical_probe}

We additionally test whether order sensitivity persists when expression identity, frequency, and model are held fixed. Expression--model pairs are grouped using only the explanation generated for the canonical input: the high-score group satisfies both $S_{Acc} \geq 0.8$ and $S_{Logic} \geq 0.8$, whereas the low-score group fails at least one criterion. We then compare Recovery on scrambled inputs by retained-local-adjacency level within each fixed expression--model pair.

\begin{table}[t]
\centering
\scriptsize
\setlength{\tabcolsep}{2.2pt}
\renewcommand{\arraystretch}{1.10}
\resizebox{\columnwidth}{!}{%
\begin{tabular}{lcccc}
\toprule
\textbf{Canonical group} & \textbf{Pairs} & \textbf{0 adj.} & \textbf{1 adj.} & \textbf{2 adj.} \\
\midrule
High-score & 27,608 & 18.4 & 28.8 & 40.2 \\
Low-score  & 14,908 &  4.8 &  7.6 & 13.1 \\
Difference & ---    & 13.6 & 21.2 & 27.1 \\
\bottomrule
\end{tabular}%
}
\caption{\textbf{Canonical-explanation probe.} Recovery (\%) on scrambled inputs by retained local adjacency.}
\label{tab:canonical_probe}
\end{table}

Recovery rises with retained local adjacency in both groups, indicating that sensitivity to internal order remains after controlling for canonical explanation quality, expression identity, frequency, and model.

\subsection{Synthetic Control Construction}
\label{app:synthetic_control}

To further examine whether reconstruction performance primarily reflects memorized idiom retrieval, we construct a synthetic control set consisting of non-attested four-character expressions.

Synthetic items are generated by sampling characters from semantically related groups (e.g., natural phenomena or seasonal terms) and combining them into four-character sequences that do not correspond to entries in standard idiom dictionaries or commonly used lexicalized expressions. Candidate sequences found in idiom dictionaries or collected lexical resources are removed. Annotators manually verify both non-attestation and compatibility with an intended compositional reading before assigning a canonical order according to the predefined compositional structure.

For each synthetic expression, a canonical order is defined and the same permutation protocol used in OrderProbe is applied. Models are evaluated using the identical reconstruction prompts and diagnostic metrics as in the main benchmark.

Table~\ref{tab:synthetic_control} reports Recovery Rate and Semantic Accuracy on idiomatic items and synthetic expressions under this setting.

\begin{table}[t]
\centering
\small
\setlength{\tabcolsep}{3.5pt}
\begin{tabular}{lcc}
\toprule
\textbf{Dataset} & \textbf{Exact (\%)} & \textbf{Sem. Fid.} \\
\midrule
Idiomatic Expr.  & 30.4 & 0.81 \\
Synthetic Expr.  & 16.8 & 0.63 \\
\bottomrule
\end{tabular}
\caption{Reconstruction performance on idiomatic expressions and synthetic non-idiomatic expressions under the same permutation protocol.}
\label{tab:synthetic_control}
\end{table}

\subsection{Reconstruction-Only Output Ablation}
\label{app:reconstruction_only}

To isolate the effect of the joint expression-plus-explanation output schema, we conduct a stratified evaluation on 200 benchmark inputs using a reconstruction-only prompt. The prompt retains the language-matched recovery instruction but requires only the recovered canonical expression. Recovery reaches 43.5\% in this condition, compared with the 20.98\% zero-shot macro-average under the standard output schema. Thus, removing the explanation requirement reduces format and generation burden, but 56.5\% of sampled inputs remain unrecovered.

\begin{table}[t]
\centering
\scriptsize
\setlength{\tabcolsep}{4pt}
\renewcommand{\arraystretch}{1.10}
\begin{tabular}{lcc}
\toprule
\textbf{Output schema} & \textbf{Inputs} & \textbf{Recovery (\%)} \\
\midrule
Standard expression + explanation & Full benchmark & 20.98 \\
Reconstruction only & 200 stratified inputs & 43.50 \\
\bottomrule
\end{tabular}
\caption{\textbf{Reconstruction-only output ablation.} The two values use different evaluation scopes and are therefore indicative comparisons.}
\label{tab:reconstruction_only}
\end{table}

\subsection{Error-Category Analysis}
\label{app:error_categories}

We characterize non-exact zero-shot outputs using a stratified sample by model, script setting, and retained-adjacency level. We sample 10 outputs from each of $12 \times 4 \times 3$ cells, yielding 1,440 outputs. Three language experts independently assign mutually exclusive categories; the Chinese panel covers both Chinese scripts. Category rates are estimated using the corresponding sampling fractions.

\begin{table}[t]
\centering
\scriptsize
\setlength{\tabcolsep}{2.5pt}
\renewcommand{\arraystretch}{1.10}
\resizebox{\columnwidth}{!}{%
\begin{tabular}{lp{4.2cm}c}
\toprule
\textbf{Category} & \textbf{Definition} & \textbf{Rate (\%)} \\
\midrule
Exact canonical recovery & Extracted expression matches the gold canonical form & 20.98 \\
Related but non-canonical & Another well-formed expression shares at least two target characters or the core meaning & 12.80 \\
Valid but unrelated & Well-formed expression without target overlap or core meaning & 8.70 \\
Literal recombination & Coherent compositional reading without recovered expression & 31.50 \\
Malformed / invalid & No unique well-formed expression or classifiable explanation & 4.60 \\
Other incorrect & Incorrect output not covered by the preceding categories & 21.42 \\
\bottomrule
\end{tabular}%
}
\caption{\textbf{Estimated error categories under zero-shot prompting.} Categories are mutually exclusive.}
\label{tab:error_categories}
\end{table}

\subsection{Frequency Under Semantic Signal Disruption}
\label{app:freq_disruption_details}
A potential concern is that exact recovery on four-character expressions may be influenced by memorization, since frequent idioms are more likely to appear in pretraining corpora.
To quantify this effect, we conduct a pilot frequency-stratified analysis under the zero-shot setting using two representative models that span a wide capability range: GPT-4o and Qwen-3-8B.
We stratify expressions into three frequency bins (high, mid, and low) and measure the recovery rate as exact canonical reconstruction.
Table~\ref{tab:freq_pilot} reports bin-wise recovery averaged across the two models.

As shown in Table~\ref{tab:freq_pilot}, we observe a clear monotonic frequency effect: recovery is highest on high-frequency expressions, decreases on mid-frequency expressions, and remains non-trivial on low-frequency expressions.
This indicates that lexical frequency contributes to reconstruction performance, while meaningful recovery persists even in the low-frequency regime.

\section{Prompt Templates for Idiom Meaning Inference}
\label{sec:appendix_prompts}

All baseline runs use a fixed two-message prompting scheme consisting of a system instruction and a user query template.
The system instruction constrains the output format to a single-line response in Chinese.
The user template provides the target idiom and repeats the required output schema to reduce formatting deviations.
Across different experimental conditions, only the prompt text is modified, while the model, decoding settings, batch processing logic, and output parsing rules remain unchanged.
For multilingual settings, the same prompting structure is retained, with the prompt content adapted through direct translation or equivalent reformulation while preserving identical functional roles and formatting requirements.

To facilitate reproducibility, Table~\ref{tab:baseline-multilingual-prompts} presents the baseline prompt templates used for idiom meaning inference across languages.
The prompts are shown in their actual formatted layout, including line breaks, to illustrate the structural constraints imposed on model outputs.
No task-specific examples, demonstrations, or additional guidance are included beyond the required format specification.

All experiments are conducted using fixed inference settings.
For the \textit{baseline} condition, no task-specific examples or chain-of-thought instructions are provided beyond the one-line format constraint.
For the \textit{CoT} condition, we use a fixed two-line CoT template (Table~\ref{tab:cot-multilingual-prompts}) while keeping all decoding settings identical.

\section{Literal Interpretation and Idiomatic Retrieval}
\label{app:case_study}

Figure~\ref{fig:case_study} presents a representative failure case on the idiom \textit{Qing Guo Qing Cheng}.
Although the scrambled characters preserve strong lexical cues, some models still produce fluent but non-canonical literal interpretations based on character-level composition.
In contrast, stronger models correctly retrieve the idiomatic metaphor, demonstrating that semantic plausibility and structural reconstruction can diverge sharply.

\begin{table*}[t!]
\centering
\footnotesize
\renewcommand{\arraystretch}{1.25}
\setlength{\tabcolsep}{10pt}

\setlength{\aboverulesep}{0pt}
\setlength{\belowrulesep}{0pt}

\definecolor{TierBlue}{RGB}{235,242,250}
\definecolor{HeaderBlue}{RGB}{245,249,253}
\definecolor{RecoverText}{RGB}{70,110,160} 

\begin{tabular}{l ccccc c}
\toprule
\rowcolor{HeaderBlue}
\textbf{Model Family} 
& $S_{Acc}^{mean}$ 
& $S_{Cons}$ 
& $S_{Logic}$ 
& $S_{Rob}$ 
& $S_{Info}$ 
& \textbf{Recovery (\%)} \\
\midrule

\multicolumn{7}{l}{
\cellcolor{TierBlue}\textbf{Tier 1: High-Performance Reasoning Models}
} \\

Qwen-3-14B                 
& 0.862 & 0.857 & 0.873 & 0.904 & 0.841 
& {\color{RecoverText}\textbf{100}} \\

Qwen-3-VLThink 
& 0.884 & 0.853 & 0.886 & 0.913 & 0.832 
& {\color{RecoverText}\textbf{100}} \\

Gemini-2.5-Flash           
& 0.871 & 0.864 & 0.859 & 0.898 & 0.821 
& {\color{RecoverText}\textbf{100}} \\

\midrule

\multicolumn{7}{l}{
\cellcolor{TierBlue}\textbf{Tier 2: Mid-Range General Models}
} \\

GPT-4o                  
& 0.879 & 0.882 & 0.858 & 0.914 & 0.843 
& {\color{RecoverText}\textbf{100}} \\

GLM-4VThink 
& 0.833 & 0.842 & 0.845 & 0.883 & 0.801 
& {\color{RecoverText}\textbf{100}} \\

DeepSeek-V3.2           
& 0.874 & 0.868 & 0.893 & 0.909 & 0.847 
& {\color{RecoverText}\textbf{100}} \\

\midrule

\multicolumn{7}{l}{
\cellcolor{TierBlue}\textbf{Tier 3: Efficient \& Baseline Models}
} \\

DeepSeek-R1        
& 0.821 & 0.848 & 0.824 & 0.879 & 0.793 
& {\color{RecoverText}\textbf{100}} \\

Claude-HaiKu-4.5   
& 0.843 & 0.861 & 0.834 & 0.887 & 0.802 
& {\color{RecoverText}\textbf{100}} \\

Gemma-3-4B         
& 0.783 & 0.819 & 0.802 & 0.847 & 0.742 
& {\color{RecoverText}\textbf{100}} \\

Llama-3-8B         
& 0.758 & 0.832 & 0.791 & 0.853 & 0.723 
& {\color{RecoverText}\textbf{100}} \\

Llama-2-7B         
& 0.742 & 0.814 & 0.774 & 0.838 & 0.703 
& {\color{RecoverText}\textbf{100}} \\

Qwen-3-8B          
& 0.779 & 0.823 & 0.788 & 0.852 & 0.739 
& {\color{RecoverText}\textbf{100}} \\

\bottomrule
\end{tabular}

\vspace{-0.2cm}
\caption{\textbf{Reference Diagnostic Metrics under Unperturbed Inputs.}
This appendix table reports diagnostic component scores obtained from a reference evaluation run using unperturbed inputs and the same inference configurations as the main experiments. Since the inputs are already in canonical order, exact reconstruction is trivially easier in this setting, and all evaluated models achieve perfect recovery on these inputs. These results are provided for contextual reference only and are not used in any comparative or statistical claims.}

\label{tab:appendix_clean_reference}
\vspace{-0.3cm}
\end{table*}

\section{Qualitative Analysis and Case Studies}
\label{sec:appendix_cases}

\begin{figure*}[p]
    \centering
    
    \includegraphics[width=1.0\linewidth, height=0.42\textheight, keepaspectratio]{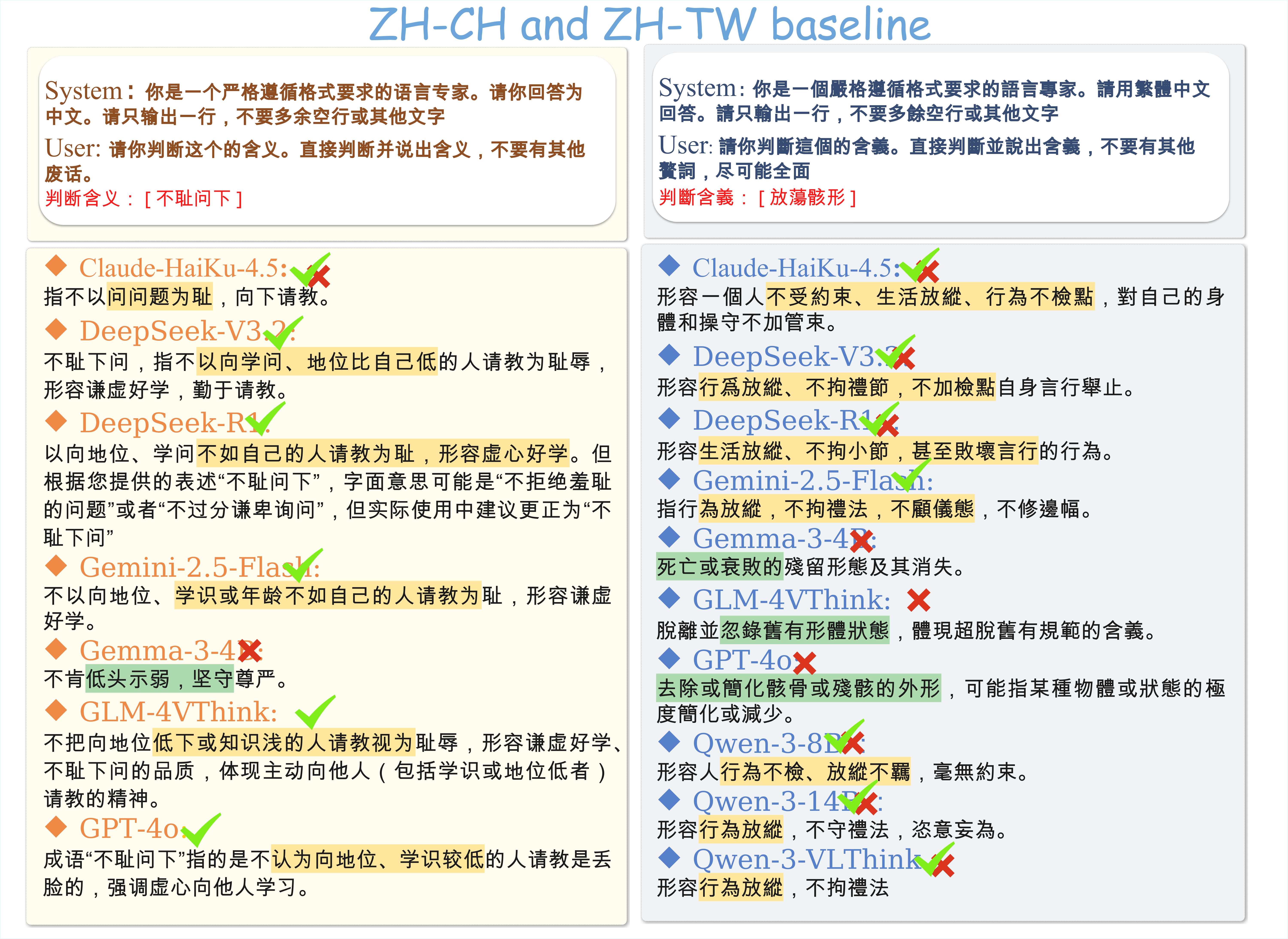}
    \caption{\textbf{Baseline performance on Chinese scripts illustrating Simplified Chinese on the left and Traditional Chinese on the right.} Even powerful models often fail to recover the canonical order.}
    \label{fig:case_baseline_zh}
    
    \vspace{1em}
    
    \includegraphics[width=1.0\linewidth, height=0.42\textheight, keepaspectratio]{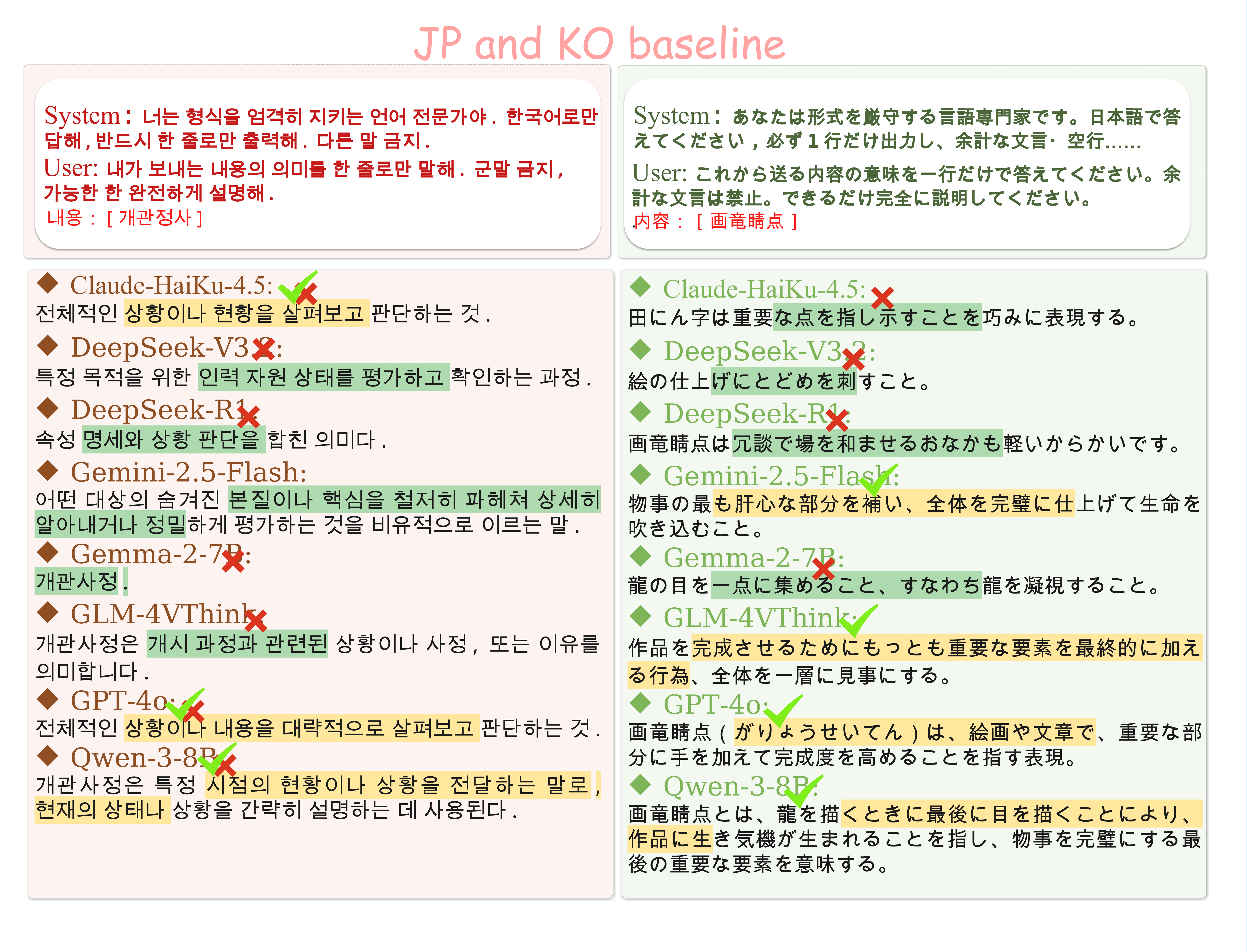}
    \caption{\textbf{Baseline performance on Japanese and Korean scripts showing Korean Hangul on the left and Japanese Kanji on the right.} The structural scrambling effectively disrupts local semantic anchors.}
    \label{fig:case_baseline_jp_ko}
\end{figure*}

\begin{figure*}[p]
    \centering
    
    \includegraphics[width=1.0\linewidth, height=0.42\textheight, keepaspectratio]{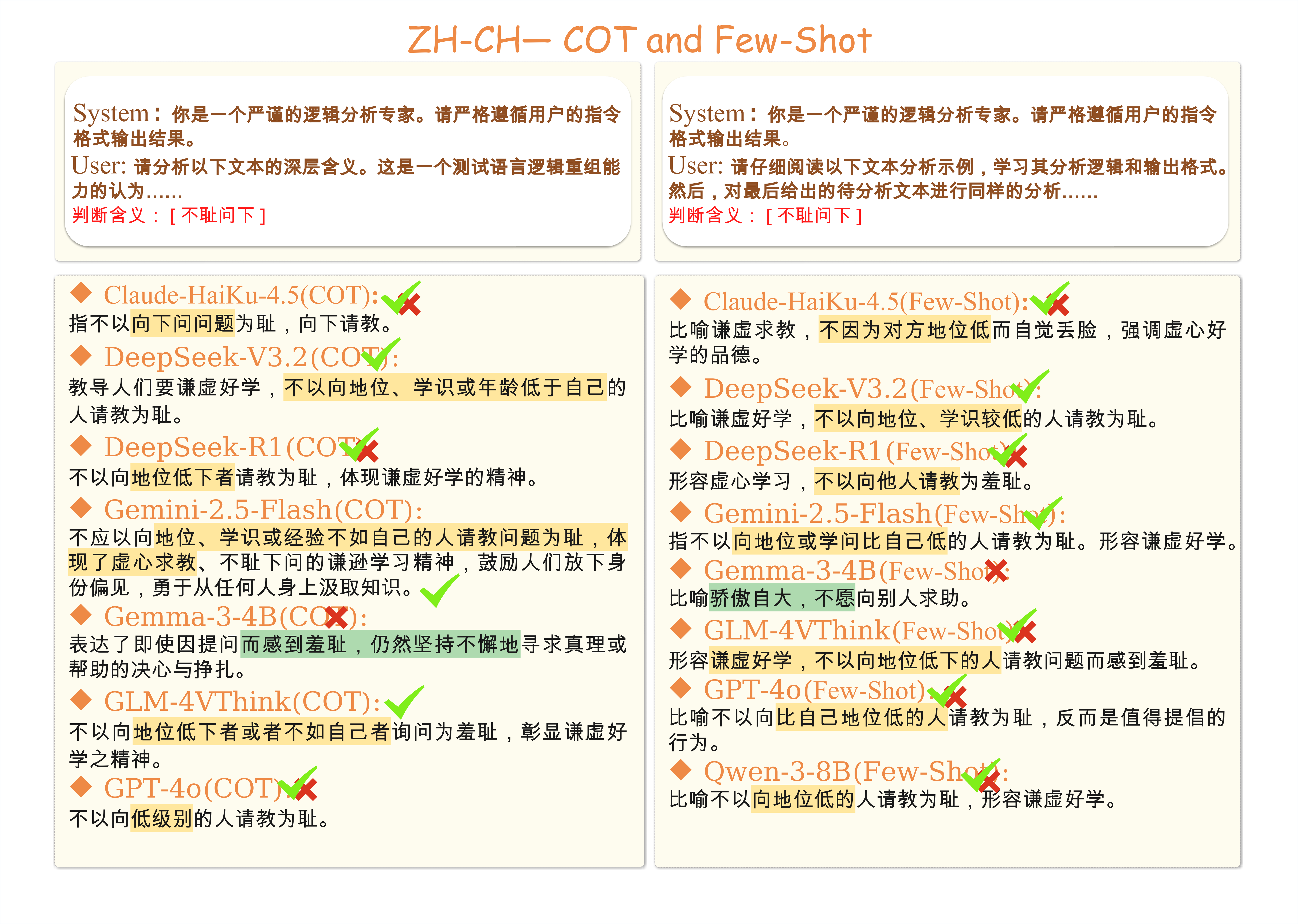}
    \caption{\textbf{Comparison of prompting strategies on Simplified Chinese.} Chain of Thought prompting elicits better reasoning but still suffers from hallucination during complex reordering.}
    \label{fig:case_cn}
    
    \vspace{1em}
    
    \includegraphics[width=1.0\linewidth, height=0.42\textheight, keepaspectratio]{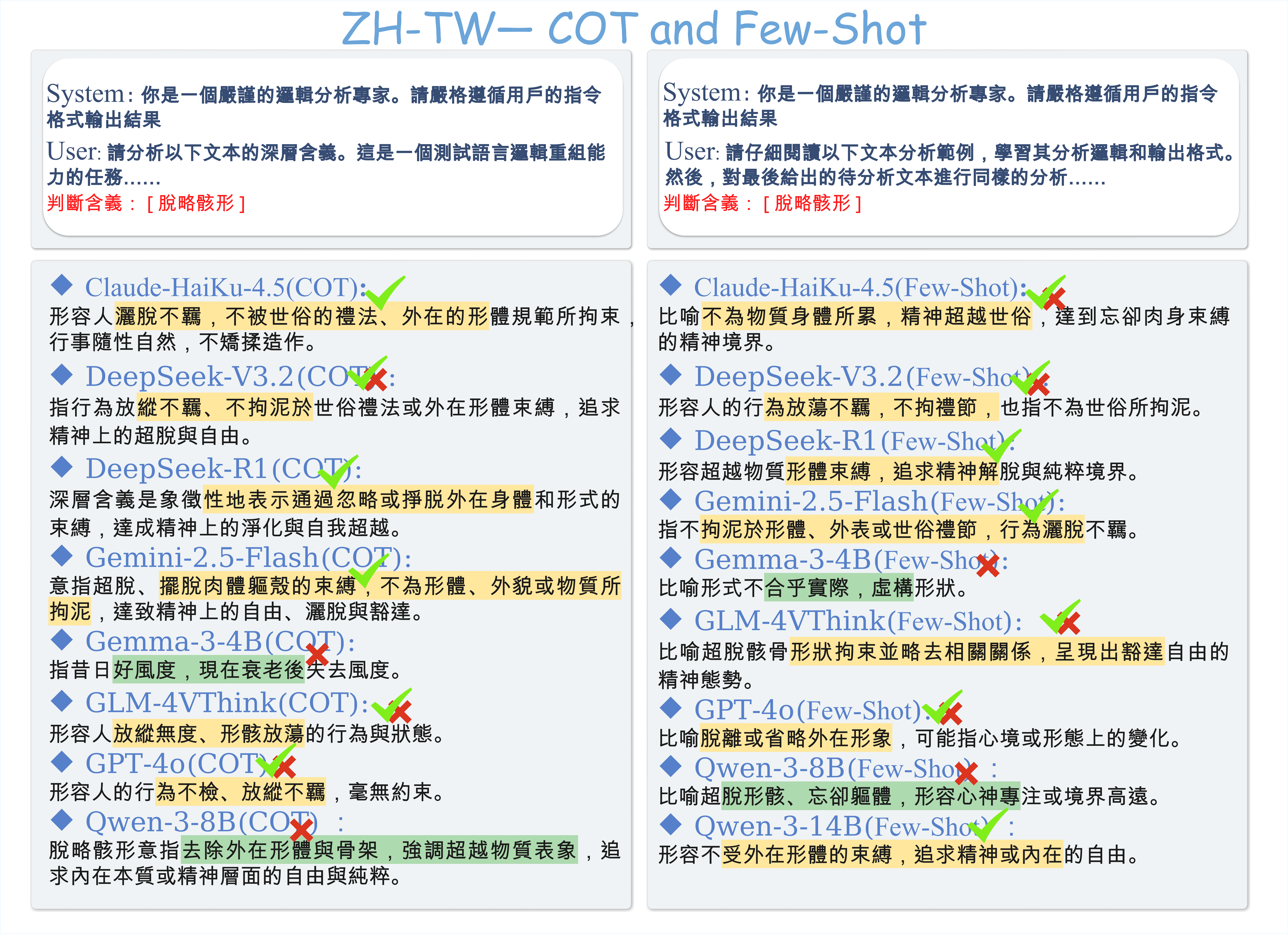}
    \caption{\textbf{Comparison of prompting strategies on Traditional Chinese.} The results highlight significant performance gaps across different inference methods.}
    \label{fig:case_tw}
\end{figure*}

\begin{figure*}[p]
    \centering
    
    \includegraphics[width=1.0\linewidth, height=0.42\textheight, keepaspectratio]{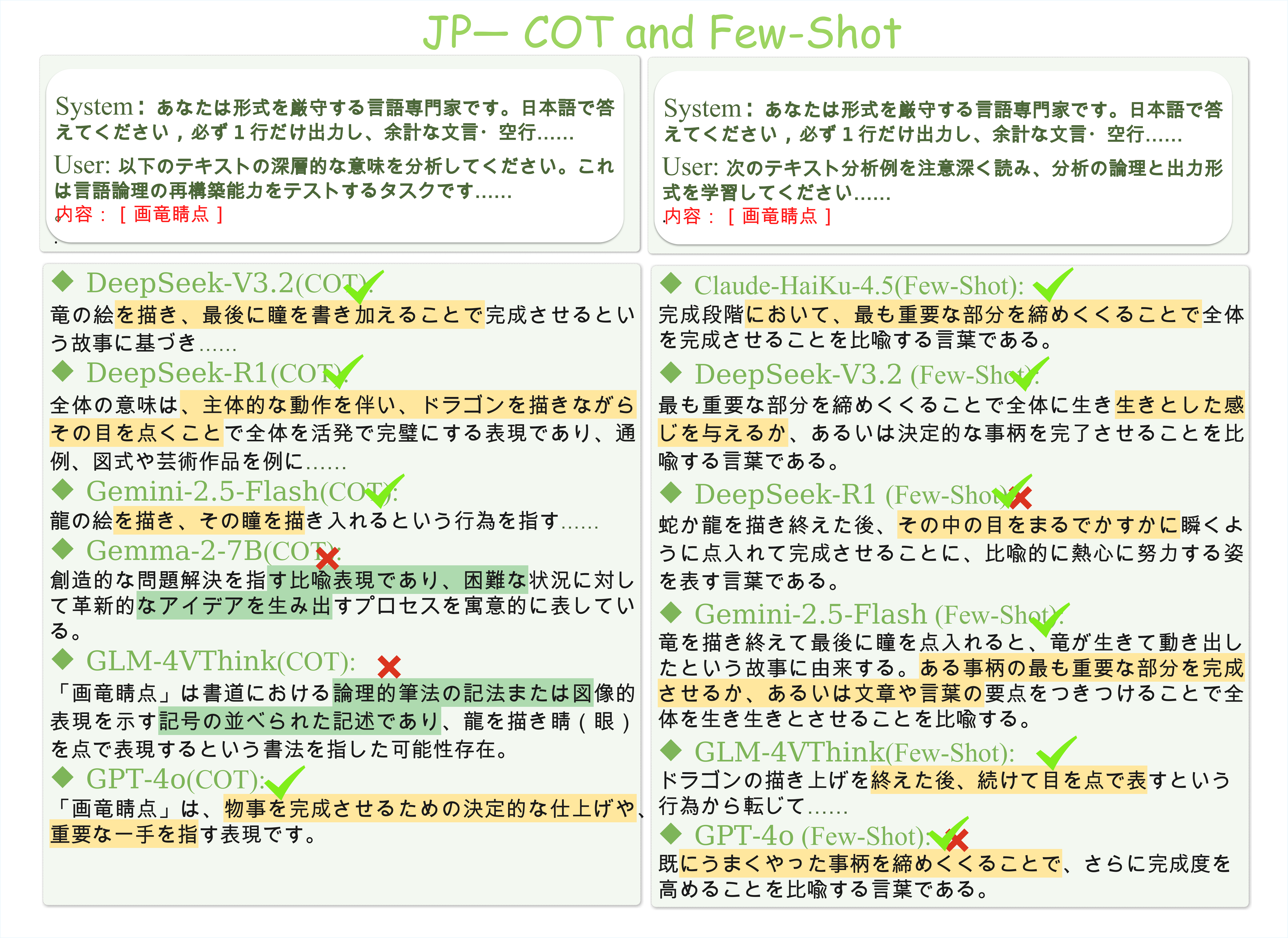}
    \caption{\textbf{Comparison of prompting strategies on Japanese.} This detailed analysis highlights the syntax recovery challenges in mixed script inputs.}
    \label{fig:case_jp}
    
    \vspace{1em}
    
    \includegraphics[width=1.0\linewidth, height=0.42\textheight, keepaspectratio]{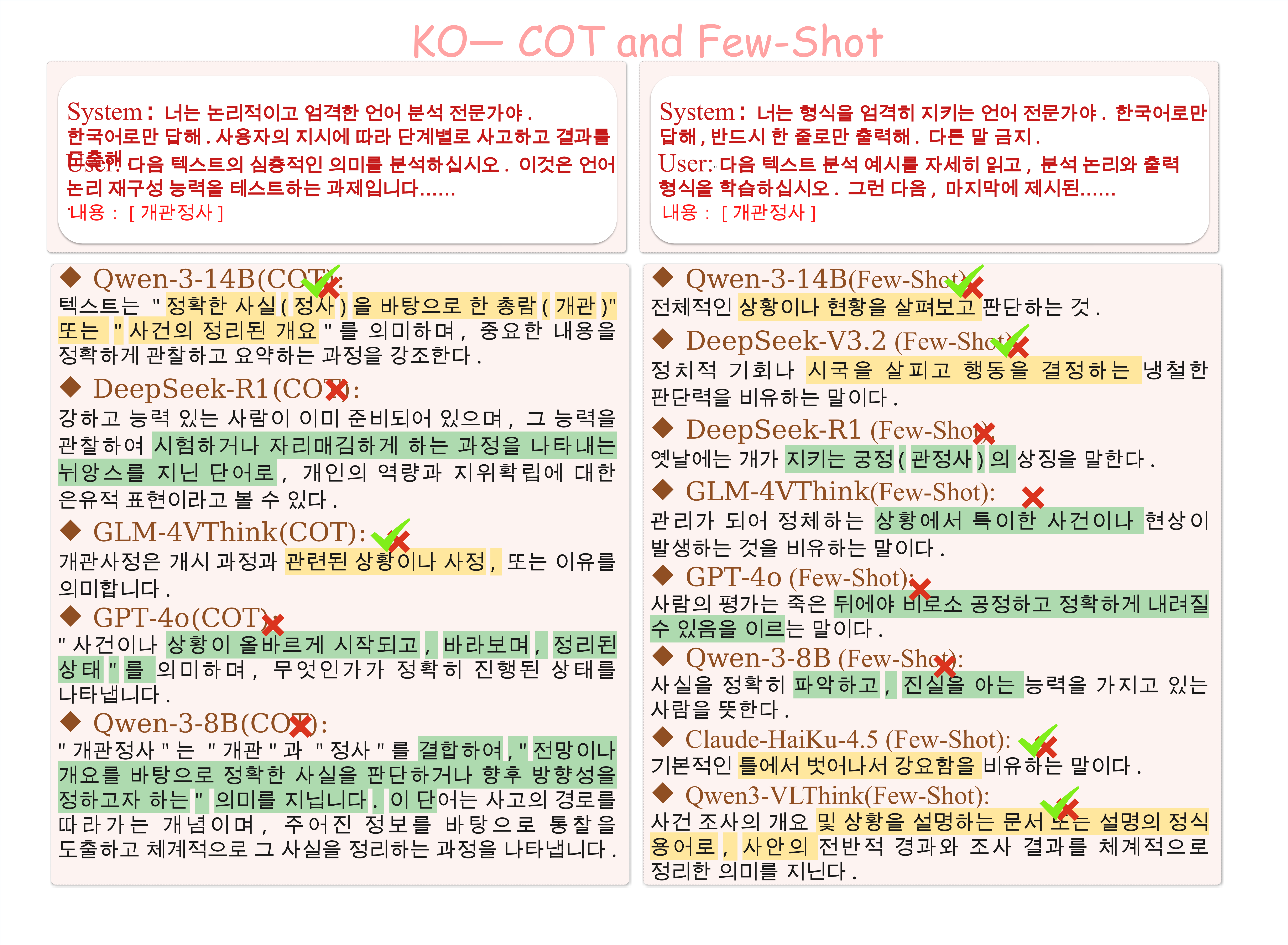}
    \caption{\textbf{Comparison of prompting strategies on Korean.} We observe distinct behaviors between standard prompting and reasoning enhanced generation.}
    \label{fig:case_ko}
\end{figure*}

In this section, we provide a qualitative visual analysis of model performance across different experimental settings.
Figures~\ref{fig:case_baseline_zh} and~\ref{fig:case_baseline_jp_ko} illustrate baseline (zero-shot) behaviors and highlight common failure modes such as hallucination and structural blindness.
Figures~\ref{fig:case_cn} through~\ref{fig:case_ko} further compare Chain-of-Thought (CoT) and few-shot strategies for each language (ZH-CN, ZH-TW, JP, KO).
These visualizations display the system prompts, user queries, and model outputs, and they illustrate the strict constraints imposed by our evaluation pipeline.

\section{Reference Performance on Canonical Inputs}
\label{sec:appendix_reference_performance}

To isolate the specific impact of structural perturbation, we conducted a control experiment using the canonical (unperturbed) forms of the expressions. Because the canonical order is already visible, this condition is a reference control for Recovery rather than a non-trivial reconstruction test.

Table \ref{tab:appendix_clean_reference} reports Recovery and diagnostic scores under this control condition. Recovery is 100\% because the canonical expression is directly available; the diagnostic scores instead evaluate the quality and logical compatibility of the model-generated explanation, and therefore need not equal 1.

These results support two limited conclusions regarding the main experiments:
\begin{enumerate}
    \item \textbf{Visible-order control:} The condition separates the difficulty of recovering the order from the difficulty of generating an explanation when the canonical input is intact.
    \item \textbf{Diagnostic separation:} The non-perfect diagnostic values are expected because they score generated explanations, not whether the model copied the visible canonical expression.
\end{enumerate}

The contrast between trivial Recovery in this control and degradation under permutation (where recovery rates frequently drop below 35\%) isolates the structural-reconstruction component discussed in the main text.

\begin{table*}[t!]
\centering
\small
\renewcommand{\arraystretch}{1.6}

\begin{tabularx}{\textwidth}{l l X}
\noalign{\hrule height 1.2pt}
\rowcolor{AclDarkBlue}
\textbf{\color{white}Language} &
\textbf{\color{white}Type} &
\textbf{\color{white}Baseline Prompt Template (Visualized Layout)} \\
\noalign{\hrule height 0.8pt}

\rowcolor{LighterBlue}
\cellcolor{white} & System &
\zh{你是一个严格遵循格式要求的语言专家。请你回答为中文。请只输出一行，不要多余空行或其他文字：}\par
\textbf{\zh{判断含义：[解释]}} \\
\rowcolor{LighterBlue}
\multirow{-3}{*}{\cellcolor{white}\textbf{ZH-CN}} & User &
\zh{请你判断这个的含义。直接判断并说出含义，不要有其他废话，尽可能全面：}\par
\textbf{\zh{判断含义：[解释]}} \par\par
\zh{成语：}\{idiom\} \\
\hline

\rowcolor{LighterGreen}
\cellcolor{white} & System &
\tw{你是一個嚴格遵循格式要求的語言專家。請用繁體中文回答。請只輸出一行，不要多餘空行或其他文字：}\par
\textbf{\tw{判斷含義：[解釋]}} \\
\rowcolor{LighterGreen}
\multirow{-3}{*}{\cellcolor{white}\textbf{ZH-TW}} & User &
\tw{請你判斷這個的含義。直接判斷並說出含義，不要有其他贅詞，尽可能全面：}\par
\textbf{\tw{判斷含義：[解釋]}} \par\par
\tw{成語：}\{idiom\} \\
\hline

\rowcolor{LighterOrange}
\cellcolor{white} & System &
\ko{너는 형식을 엄격히 지키는 언어 전문가다. 한국어로만 답하라. 반드시 한 줄로만 출력하고 다른 문장은 포함하지 마라：}\par
\textbf{\ko{의미: [설명]}} \\
\rowcolor{LighterOrange}
\multirow{-3}{*}{\cellcolor{white}\textbf{KO}} & User &
\ko{다음에 제시된 표현의 의미를 한 줄로 설명하라. 군말은 포함하지 말고 가능한 한 완전하게 설명하라：}\par
\textbf{\ko{의미: [설명]}} \par\par
\ko{내용：}\{idiom\} \\
\hline

\rowcolor{LighterPurple}
\cellcolor{white} & System &
\ja{あなたは形式を厳守する言語専門家です。日本語で回答してください。必ず一行のみを出力し余分な文言や空行は含めないでください：}\par
\textbf{\ja{意味：[説明]}} \\
\rowcolor{LighterPurple}
\multirow{-3}{*}{\cellcolor{white}\textbf{JA}} & User &
\ja{以下に示す表現の意味を一行で説明してください。余分な文言は禁止し可能な限り完全に述べてください：}\par
\textbf{\ja{意味：[説明]}} \par\par
\ja{内容：}\{idiom\} \\

\noalign{\hrule height 1.2pt}
\end{tabularx}

\caption{Visualized baseline prompt templates used for idiom meaning inference across four languages. Prompts are shown in their formatted layout to illustrate the structural constraints imposed on model outputs.}
\label{tab:baseline-multilingual-prompts}
\end{table*}

\subsection{English Readings of Prompt Templates}
\label{app:prompt_translations}

For accessibility, the following English readings accompany the exact multilingual prompts above and the few-shot/CoT templates below. They preserve the operational constraints (language, output field, and line count); the original-language strings remain the prompts used for evaluation.

\begin{center}
\centering
\scriptsize
\setlength{\tabcolsep}{3.5pt}
\renewcommand{\arraystretch}{1.12}
\begin{tabularx}{\linewidth}{l X}
\toprule
\textbf{Prompt block} & \textbf{English reading} \\
\midrule
Baseline system & You are a language expert who follows the format strictly. Answer in the target language and output exactly one line: ``Meaning: [explanation].'' \\
Baseline user & Explain the meaning of the expression below as completely as possible, without filler: ``Expression: \{idiom\}.'' \\
Few-shot system & You are a rigorous logic-analysis expert. Follow the requested format and do not add an opening or closing remark. \\
Few-shot user & Read the analysis examples, learn their logic and format, and apply the same two-line analysis to the final expression. \\
CoT user & Analyze the expression step by step: decompose character meanings, reconstruct the likely order, and derive the overall meaning. Output exactly two lines: ``Thinking Process: ...'' and ``Final Meaning: ...''. \\
\bottomrule
\end{tabularx}
\captionof{table}{Functional English readings of the multilingual prompt templates. Language-specific wording (including the Korean and Japanese labels) is retained in the exact templates.}
\label{tab:prompt_english_readings}
\end{center}

\begin{table*}[t!]
\centering
\tiny
\renewcommand{\arraystretch}{1.05}
\setlength{\tabcolsep}{3pt}
\captionsetup{font=footnotesize,skip=2pt}

\begin{tabularx}{\textwidth}{l l >{\raggedright\arraybackslash}X}
\noalign{\hrule height 1.2pt}
\rowcolor{AclDarkBlue}
\color{white}Language &
\color{white}Type &
\color{white}Few Shot Prompt Template Visualized Layout \\
\noalign{\hrule height 0.8pt}

\rowcolor{LighterBlue}
\cellcolor{white}ZH-CN & System &
\zh{你是一个严谨的逻辑分析专家。请严格遵循用户的指令格式输出结果。禁止输出任何开场白或结束语。} \\

\rowcolor{LighterBlue}
\cellcolor{white} & System &
start deep thinking, please always use the thinking mode \\

\rowcolor{LighterBlue}
\cellcolor{white} & User &
\zh{请仔细阅读以下文本分析示例，学习其分析逻辑和输出格式。然后，对最后给出的待分析文本进行同样的分析。}\par
\zh{示例 1}\par
\zh{输入文本：惊人一鸣}\par
\zh{Thinking Process：分析字面，惊人一鸣在语义上不通顺。观察字词组合，发现一鸣通常作为主语或状语，惊人作为结果。调整语序为一鸣惊人，符合一声鸣叫让人震惊的逻辑。}\par
\zh{Final Meaning：比喻平时没有突出的表现，一下子做出惊人的成绩。}\par
\zh{示例 2}\par
\zh{输入文本：添足画蛇}\par
\zh{Thinking Process：分析字面，添足意味着增加脚，画蛇是动作主体。逻辑上先有画蛇，后有添足的多余行为。重组为画蛇添足。}\par
\zh{Final Meaning：比喻做了多余的事，非但无益，反而不合适。}\par
\zh{示例 3}\par
\zh{输入文本：求剑刻舟}\par
\zh{Thinking Process：分析字面，求剑是目的，刻舟是手段。按照因果和动作发生的时间顺序，应先在船上刻记号，再去寻找剑。重组为刻舟求剑。}\par
\zh{Final Meaning：比喻拘泥不知变通，不懂得根据实际情况处理问题。}\par
\zh{待分析文本}\par
\zh{输入文本：}\{idiom\}\par
\zh{请严格模仿上述示例的逻辑和格式输出，必须严格两行。} \\
\hline

\rowcolor{LighterGreen}
\cellcolor{white}ZH-TW & System &
\tw{你是一個嚴謹的邏輯分析專家。請嚴格遵循用戶的指令格式輸出結果。禁止輸出任何開場白或結束語。} \\

\rowcolor{LighterGreen}
\cellcolor{white} & System &
start deep thinking, please always use the thinking mode \\

\rowcolor{LighterGreen}
\cellcolor{white} & User &
\tw{請仔細閱讀以下文本分析範例，學習其分析邏輯和輸出格式。然後，對最後給出的待分析文本進行同樣的分析。}\par
\tw{範例 1}\par
\tw{輸入文本：驚人一鳴}\par
\tw{Thinking Process：分析字面，驚人一鳴在語義上不通順。觀察字詞組合，發現一鳴通常作為主語或狀語，驚人作為結果。調整語序為一鳴驚人，符合一聲鳴叫讓人震驚的邏輯。}\par
\tw{Final Meaning：比喻平時沒有突出的表現，一下子做出驚人的成績。}\par
\tw{範例 2}\par
\tw{輸入文本：添足畫蛇}\par
\tw{Thinking Process：分析字面，添足意味著增加腳，畫蛇是動作主體。邏輯上先有畫蛇，後有添足的多餘行為。重組為畫蛇添足。}\par
\tw{Final Meaning：比喻做了多餘的事，非但無益，反而不合適。}\par
\tw{範例 3}\par
\tw{輸入文本：求劍刻舟}\par
\tw{Thinking Process：分析字面，求劍是目的，刻舟是手段。按照因果和動作發生的時間順序，應先在船上刻記號，再去尋找劍。重組為刻舟求劍。}\par
\tw{Final Meaning：比喻拘泥不知變通，不懂得根據實際情況處理問題。}\par
\tw{待分析文本}\par
\tw{輸入文本：}\{idiom\}\par
\tw{請嚴格模仿上述範例的邏輯和格式輸出，必須嚴格兩行。} \\
\hline

\rowcolor{LighterOrange}
\cellcolor{white}KO & System &
\ko{너는 논리적이고 엄격한 언어 분석 전문가야. 한국어로만 답해. 사용자의 지시에 따라 단계별로 사고하고 결과를 도출해.} \\

\rowcolor{LighterOrange}
\cellcolor{white} & System &
start deep thinking, please always use the thinking mode \\

\rowcolor{LighterOrange}
\cellcolor{white} & User &
\ko{다음 텍스트 분석 예시를 자세히 읽고 분석 논리와 출력 형식을 학습하십시오. 그런 다음 마지막에 제시된 분석 대상 텍스트에 대해 동일한 분석을 수행하십시오.}\par
\ko{예시 1}\par
\ko{입력 텍스트: 경인일명 (驚人一鳴)}\par
\ko{Thinking Process: 새가 한 번 울어 사람들을 놀라게 한다는 인과 관계가 자연스럽다. 따라서 어순을 일명경인으로 재구성한다.}\par
\ko{Final Meaning: 평소에는 조용하다가 갑자기 놀라운 성과를 내어 세상을 놀라게 함을 비유하는 말이다.}\par
\ko{예시 2}\par
\ko{입력 텍스트: 첨족화사 (添足畫蛇)}\par
\ko{Thinking Process: 뱀을 그리고 나서 발을 더하는 행위가 불필요하다. 어순을 화사첨족으로 재구성한다.}\par
\ko{Final Meaning: 쓸데없는 일을 덧붙여서 도리어 일을 그르침을 이르는 말이다.}\par
\ko{예시 3}\par
\ko{입력 텍스트: 구검각주 (求劍刻舟)}\par
\ko{Thinking Process: 칼을 찾기 위한 수단으로 배에 표시를 하므로 먼저 각주하고 나중에 구검한다. 어순을 각주구검으로 재구성한다.}\par
\ko{Final Meaning: 변화하는 상황을 모르고 낡은 방식에 집착하는 어리석음을 비유한다.}\par
\ko{분석 대상 텍스트}\par
\ko{입력 텍스트: }\{idiom\}\par
\ko{위 예시들의 논리와 형식을 엄격히 모방하여 출력하십시오. 반드시 두 줄만 출력하십시오.} \\
\hline

\rowcolor{LighterPurple}
\cellcolor{white}JA & System &
\ja{あなたは厳格な論理を持つ言語分析の専門家です。ユーザーの指示するフォーマットに厳密に従い、余計な挨拶や結びの言葉は出力しないでください。} \\

\rowcolor{LighterPurple}
\cellcolor{white} & System &
start deep thinking, please always use the thinking mode \\

\rowcolor{LighterPurple}
\cellcolor{white} & User &
\ja{次のテキスト分析例を注意深く読み、分析の論理と出力形式を学習してください。その後、最後に示される分析対象テキストについて同様の分析を行ってください。}\par
\ja{例 1}\par
\ja{入力テキスト：驚人一鳴}\par
\ja{Thinking Process: 驚人一鳴は不自然であり、一鳴が原因で驚人が結果となるため語順を一鳴驚人に調整する。}\par
\ja{Final Meaning: 普段は目立たないが突然すばらしい成果を出して人々を驚かせることのたとえ。}\par
\ja{例 2}\par
\ja{入力テキスト：添足画蛇}\par
\ja{Thinking Process: まず蛇を描き、その後で足を足す行為が余計である。語順を画蛇添足に再構成する。}\par
\ja{Final Meaning: 余計なことをしてかえって不適切になることのたとえ。}\par
\ja{例 3}\par
\ja{入力テキスト：求剣刻舟}\par
\ja{Thinking Process: 剣を探すために舟に印を刻すので先に刻舟し後に求剣する。語順を刻舟求剣に再構成する。}\par
\ja{Final Meaning: 状況の変化を理解せず古い方法に固執する愚かさのたとえ。}\par
\ja{分析対象テキスト}\par
\ja{入力テキスト：}\{idiom\}\par
\ja{上記の例の論理と形式を厳密に模倣して出力してください。必ず二行のみを出力してください。} \\

\noalign{\hrule height 1.2pt}
\end{tabularx}

\caption{Visualized few shot prompt templates used for multilingual idiom analysis. Prompts are shown in formatted layout to illustrate in context reasoning patterns and the strict two line output constraint.}
\label{tab:fewshot-multilingual-prompts}

\end{table*}

\begin{table*}[p]
\centering
\footnotesize
\renewcommand{\arraystretch}{1.12}
\setlength{\tabcolsep}{5pt}
\captionsetup{font=footnotesize,skip=4pt}

\begin{tabularx}{\textwidth}{l l >{\raggedright\arraybackslash}X}
\noalign{\hrule height 1.2pt}
\rowcolor{AclDarkBlue}
\color{white}Language &
\color{white}Type &
\color{white}CoT Prompt Template Visualized Layout \\
\noalign{\hrule height 0.8pt}

\rowcolor{LighterBlue}
\cellcolor{white}ZH-CN & System &
\zh{你是一个严谨的逻辑分析专家。请严格遵循用户的指令格式输出结果。禁止输出任何开场白或结束语。} 
\par
start deep thinking, please always use the thinking mode \\

\rowcolor{LighterBlue}
\cellcolor{white} & User &
\zh{请分析以下文本的深层含义。这是一个测试语言逻辑重组能力的任务。}\par
\zh{为了确保准确性，请务必一步步思考，严格遵循以下分析路径。}\par
\zh{1 字义拆解 分析文本中每个字的独立含义。}\par
\zh{2 结构重组 观察这些字之间可能存在的句法逻辑，判断是否存在语序错位。}\par
\zh{3 语义推导 基于重组后的逻辑，推导该文本的整体寓意。}\par
\zh{请严格按照以下特定格式输出，必须严格两行。}\par
\zh{Thinking Process: 在这里写下你的详细分步推理过程，包括对字序和结构的分析。}\par
\zh{Final Meaning: 基于推理得出的最终含义解释，只给结论，不要再解释推理过程，尽可能全面。}\par
\zh{输入文本：}\{idiom\} \\[-2pt]
\noalign{\vskip 3pt}\hline\noalign{\vskip 3pt}

\rowcolor{LighterGreen}
\cellcolor{white}ZH-TW & System &
\tw{你是一個嚴謹的邏輯分析專家。請嚴格遵循用戶的指令格式輸出結果。禁止輸出任何開場白或結束語。}
\par
start deep thinking, please always use the thinking mode \\

\rowcolor{LighterGreen}
\cellcolor{white} & User &
\tw{請分析以下文本的深層含義。這是一個測試語言邏輯重組能力的任務。}\par
\tw{為了確保準確性，請務必一步步思考，嚴格遵循以下分析路徑。}\par
\tw{1 字義拆解 分析文本中每個字的獨立含義。}\par
\tw{2 結構重組 觀察這些字之間可能存在的句法邏輯，判斷是否存在語序錯位。}\par
\tw{3 語義推導 基於重組後的邏輯，推導該文本的實際整體寓意。}\par
\tw{請嚴格按照以下特定格式輸出，必須嚴格兩行。}\par
\tw{Thinking Process: 在這裡寫下你的詳細分步推理過程，包括對字序和結構的分析。}\par
\tw{Final Meaning: 基於推理得出的最終含義解釋，只給結論，不要再解釋推理過程，盡可能全面。}\par
\tw{輸入文本：}\{idiom\} \\[-2pt]
\noalign{\vskip 3pt}\hline\noalign{\vskip 3pt}

\rowcolor{LighterOrange}
\cellcolor{white}KO & System &
\ko{너는 논리적이고 엄격한 언어 분석 전문가야. 한국어로만 답해. 사용자의 지시에 따라 단계별로 사고하고 결과를 도출해.}
\par
start deep thinking, please always use the thinking mode \\

\rowcolor{LighterOrange}
\cellcolor{white} & User &
\ko{다음 텍스트의 심층적인 의미를 분석하십시오. 이것은 언어 논리 재구성 능력을 테스트하는 과제입니다.}\par
\ko{정확성을 보장하기 위해 반드시 단계별로 사고하고 다음 분석 경로를 엄격히 따르십시오.}\par
\ko{1 글자 풀이 텍스트에 포함된 각 글자의 개별적인 의미를 분석합니다.}\par
\ko{2 구조 재구성 글자들 간의 잠재적인 구문 논리를 관찰하여 어순이 뒤섞여 있는지 판단합니다.}\par
\ko{3 의미 도출 재구성된 논리를 바탕으로 텍스트의 전체 의미를 추론합니다.}\par
\ko{다음 형식에 맞춰 엄격히 출력하십시오. 반드시 두 줄이어야 하며 줄을 추가하거나 빼지 마십시오.}\par
\ko{Thinking Process: 여기에 글자 순서와 구조 분석을 포함한 상세한 단계별 추론 과정을 작성하십시오.}\par
\ko{Final Meaning: 추론을 통해 도출된 최종 의미 설명. 결론만 제시하고 추론 과정을 반복하지 마십시오. 최대한 포괄적으로 작성하십시오.}\par
\ko{입력 텍스트: }\{idiom\} \\[-2pt]
\noalign{\vskip 3pt}\hline\noalign{\vskip 3pt}

\rowcolor{LighterPurple}
\cellcolor{white}JA & System &
\ja{あなたは厳格な論理を持つ言語分析の専門家です。ユーザーの指示するフォーマットに厳密に従い、余計な挨拶や結びの言葉は出力しないでください。}
\par
start deep thinking, please always use the thinking mode \\

\rowcolor{LighterPurple}
\cellcolor{white} & User &
\ja{以下のテキストの深層的な意味を分析してください。これは言語論理の再構築能力をテストするタスクです。}\par
\ja{正確性を確保するため必ずステップバイステップで思考し以下の分析パスを厳守してください。}\par
\ja{1 文字の解読 テキストに含まれる各文字の個別の意味を分析します。}\par
\ja{2 構造の再構築 文字間の構文論理を観察し語順が入れ替わっていないか判断します。}\par
\ja{3 意味の導出 再構築された論理に基づいてテキスト全体の意味を推論します。}\par
\ja{以下の形式に厳密に従って出力してください。必ず二行でなければなりません。}\par
\ja{Thinking Process: ここに文字の順序と構造の分析を含む詳細な段階的推論プロセスを記述してください。}\par
\ja{Final Meaning: 推論に基づいて導き出された最終的な意味の説明。結論のみを提示し推論過程を繰り返さないでください。可能な限り包括的に記述してください。}\par
\ja{入力テキスト：}\{idiom\} \\

\noalign{\hrule height 1.2pt}
\end{tabularx}

\caption{Visualized CoT prompt templates used for multilingual idiom analysis. The prompts enforce a fixed two line output format and instruct stepwise reasoning for semantic and structural reconstruction.}
\label{tab:cot-multilingual-prompts}
\end{table*}

\end{document}